%% file: main.tex
\documentclass[10pt,twocolumn,letterpaper]{article}

\usepackage[pagenumbers]{iccv} % To force page numbers, e.g. for an arXiv version


\usepackage{booktabs}       % professional-quality tables
\usepackage{amsfonts}       % blackboard math symbols
\usepackage{nicefrac}       % compact symbols for 1/2, etc.
\usepackage{microtype}      % microtypography
\usepackage{xcolor}         % colors
\usepackage{graphicx} 
\usepackage{amsmath}
\usepackage{adjustbox}
\usepackage{multirow}
\usepackage{enumitem}
\usepackage{arydshln}
\usepackage{marvosym} 
\usepackage[accsupp]{axessibility}
\usepackage[toc,page]{appendix}
\makeatletter

\def\@fnsymbol#1{%
  \ifcase#1\or
    *\relax              % 1 -> 星号（等贡献脚注）
  \or
    \Letter\relax        % 2 -> 信封（对应作者脚注）
  \else
    \@ctrerr             % 超过两个脚注就报错
  \fi
}
\makeatother
\definecolor{iccvblue}{rgb}{0.21,0.49,0.74}
\usepackage[pagebackref,breaklinks,colorlinks,allcolors=iccvblue]{hyperref}

\def\paperID{6992} % *** Enter the Paper ID here
\def\confName{ICCV}
\def\confYear{2025}

\title{Pretrained Reversible Generation as \\ Unsupervised Visual Representation Learning}

\author{
Rongkun Xue$^{\spadesuit}$\thanks{Equal contribution.},
Jinouwen Zhang$^{\heartsuit}$\footnotemark[1], 
Yazhe Niu$^{\heartsuit\clubsuit}$, 
Dazhong Shen$^{\diamondsuit}$, 
Bingqi Ma$^{\blacklozenge}$, 
Yu Liu$^{\blacklozenge}$, 
Jing Yang$^{\spadesuit}$\thanks{Corresponding author.} \\
$^{\spadesuit}$Xi'an Jiaotong University
$^{\heartsuit}$Shanghai AI Laboratory
$^{\blacklozenge}$SenseTime\\
$^{\clubsuit}$The Chinese University of Hong Kong
$^{\diamondsuit}$Nanjing University of Aeronautics and Astronautics\\
{\tt\small
xuerongkun@stu.xjtu.edu.cn, 
zhangjinouwen@pjlab.org.cn, 
niuyazhe314@outlook.com
}\\
}

\begin{document}
\maketitle
\input{sec/0_abstract}
\input{sec/1_intro}
\input{sec/3_related_work}
\input{sec/4_method}
\input{sec/5_exp}

\input{sec/6_conclusion}
\clearpage
\paragraph{Acknowledgments.}
{This work was supported by the Shanghai Artificial Intelligence Laboratory, in part by the National Key R\&D Program of China under Grant No.~2022ZD0119301, and in part by the National Natural Science Foundation of China under Grant No.~U21A20485 and No.~62406141. The Work was done during Rongkun Xue’s internship at Shanghai AI Laboratory.
{
    \small
    \bibliographystyle{ieeenat_fullname}
    \bibliography{main}
}
\newpage
\input{sec/X_suppl}

\end{document}

%% file: sec/0_abstract.tex
\begin{abstract}
Recent generative models based on score matching and flow matching have significantly advanced generation tasks, but their potential in discriminative tasks remains underexplored. Previous approaches, such as generative classifiers, have not fully leveraged the capabilities of these models for discriminative tasks due to their intricate designs.
We propose Pretrained Reversible Generation (PRG), which extracts unsupervised representations by reversing the generative process of a pretrained continuous generation model.
PRG effectively reuses unsupervised generative models, leveraging their high capacity to serve as robust and generalizable feature extractors for downstream tasks.
This framework enables the flexible selection of feature hierarchies tailored to specific downstream tasks.
Our method consistently outperforms prior approaches across multiple benchmarks, achieving state-of-the-art performance among generative model based methods, including 78\% top-1 accuracy on ImageNet at a resolution of 64×64. Extensive ablation studies, including out-of-distribution evaluations, further validate the effectiveness of our approach. PRG is available at \url{https://opendilab.github.io/PRG/}.
\end{abstract}

%% file: sec/1_intro.tex
%\vspace{-12pt}
\section{Introduction}
%\vspace{-4pt}
\label{sec:intro}

\begin{figure}[h]
    \small
    \centering
      \includegraphics[width=\linewidth]{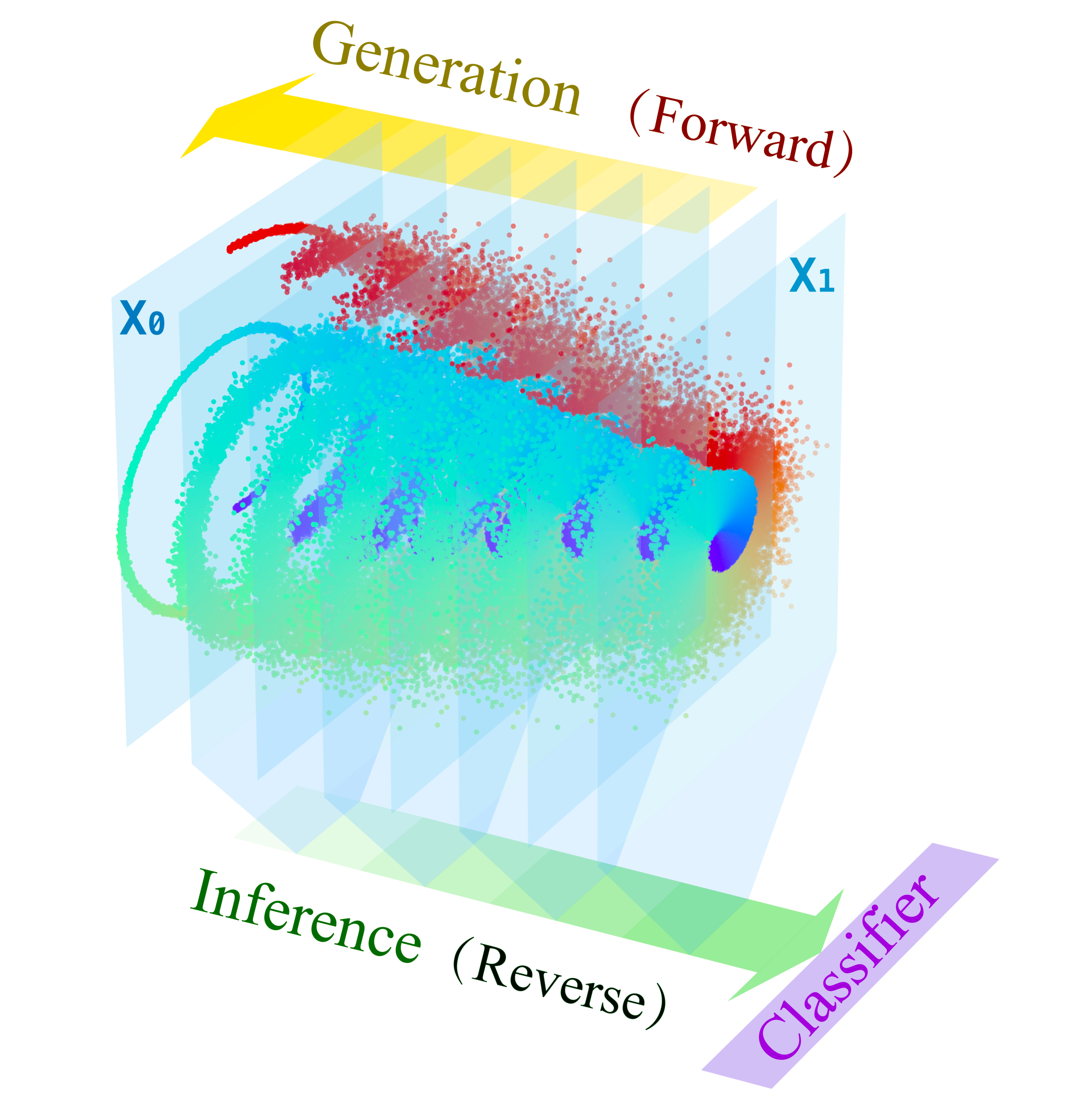}
      \caption{Overview of the PRG as Unsupervised Visual Representation pipeline. Swiss-roll data is generated using $(x, y) = \left( t \cos(t), t \sin(t) \right)$ with $t \in [0, 3\pi]$, where the color gradient from blue to red corresponds to $t$ increasing from $0$ to $3\pi$. 
      The yellow trajectory represents the generative (forward) process, with color intensity indicating direction. Green arrows illustrate the inference (reverse) generation process. Using the pretrained model, each step along the green path can be fine-tuned for downstream tasks.
      %After training the model, we apply the reverse operation.
      %Data points of different colors separate from each other, 
      %while those with the same color become thoroughly mixed. 
      %In the diffusion model, the process transitions from  $x_1$  to  $x_0$ ; 
      %in contrast, in the flow model, it moves from  $x_0$  to  $x_1$ .  
      }
      \label{fig:Pipeline of reverse process}
      %\vspace{-18pt}
  \end{figure}
  
% 紧跟在 \section{Introduction} 之后
\begin{center}
  \begin{minipage}{\linewidth}   % 0.8 视需要可改
    \centering\itshape
    ``What I cannot create, I do not understand.''
    
    %\vspace{0.4em}                  % 控制引句与署名距离
    \raggedleft—\,Richard Feynman   % 不加下划线
  \end{minipage}
\end{center}
\vspace{-0.5\baselineskip}   
Generative models formulated as continuous-time stochastic differential equations, including diffusion and flow models \citep{lipman2023flow, liu2022flow, pmlr-v202-pooladian23a, Albergo2023, shen2024rethinking}, 
have shown remarkable proficiency in multi-modal content generation tasks \citep{rombach2022high, Ho2022video, mittal2021symbolic,wang2024phased}, as well as in scientific modeling \citep{abramson2024accurate,zhang2024generative_rl}.
By effectively learning high-dimensional distributions, these models can generate new, high-quality samples that resemble the original data.

% While generative tasks require machine learning models to reconstruct data from random noise, which is a challenging task, discriminative tasks, such as image classification, 
% involve learning representations that capture the underlying data structure and classifying data based on these learned representations. 
% Previous works on representation learning rely on methods such as contrastive learning, 
% which learns desirable representations using generative models like GANs and VAEs, where encoders are naturally incorporated into their designs 
While generative tasks require models to reconstruct the entire data distribution, discriminative tasks (e.g., image classification) rely on learning representations where some parts capture the essential underlying data structure to produce discrete or continuous predictions.
Previous works~\cite{dosovitskiy2020image,he2016deep} have explored various approaches to enhance representation learning, with growing focus on unsupervised visual representation learning~\cite{chen2020simple,chen2020generative}. These methods construct generic proxy tasks to efficiently extract and utilize data for pre-training, yielding robust representations for diverse downstream tasks~\citep{bengio2013representation,radford2015unsupervised,chen2016infogan,van2017neural,donahue2019large,chen2020generative,xiang2023denoising}.

Due to the extraordinary performance of diffusion models in generative tasks, recent studies have explored their potential for discriminative tasks~\citep{clark2024text,li2023your}. 
These studies often rely on extracting or combining representations from the internal layers of pretrained diffusion models, involving intricate and poorly generalizable designs, yet usually lacking a clear explanation. For example, \cite{xiang2023denoising} selects a specific layer in a UNet-based diffusion model as the feature representation.
Moreover, the performance of these methods exhibits significant room for improvement compared to current state-of-the-art techniques.

Inspired by the reversible property~\cite{Chen2018,song2020score} of continuous-time stochastic flow models, we introduce Pretrained Reversible Generation as Unsupervised Visual Representation Learning (PRG).
Our approach applies generative models to discriminative tasks by reversing the generative process into a meaningful inference process.
Specifically, we pretrain a reversible generative model via flow matching to maximize the lower bound of mutual information between the original image and its optimal representation, providing a strong initialization. We then reverse the generative process and leverage it for downstream training to improve task performance.
This model-agnostic method does not require access to any internal features of a pretrained generative model, eliminating the need for fixed network modules to capture data representations. As a pretrain-finetune paradigm, it fully leverages the powerful capabilities of various pretrained generative models (e.g., diffusion models, flow models). During the reversed process, it utilizes features from different hierarchical levels, enabling efficient adaptation to downstream tasks.
Through theoretical analysis~\cref{sec:theroy} and verification experiments~\cref{sec:Verification Analysis}, we explain why reversing a pretrained generative process serves as an effective feature extractor for downstream tasks. Our main experiments further demonstrate that this method achieves competitive performance in image classification and OOD detection while effectively adapting community-pretrained text-to-image models for downstream training.

Our contributions can be summarized as follows:
\begin{itemize}[left=0pt]
  \item We propose a general, model-agnostic method for applying pretrained generative models to discriminative tasks.
  \item We analyze the theoretical foundation of the method and provide empirical findings demonstrating its efficacy.
  \item Extensive experiments demonstrate the robustness and generalizability of PRG in many image classification datasets.
\end{itemize}

%% file: sec/3_related_work.tex
\section{Related Work}
\label{sec:related}

\paragraph{Generative versus Discriminative Classifiers}
Generative classifiers, first proposed in \citep{li2023your, chen2023robust, clark2024text, chendiffusion, zimmermann2021score, pmlr-v162-nie22a}, model the conditional distribution $p(x\mid y)$ and use Bayes’ rule to compute $p(y\mid x)=\frac{p(x\mid y),p(y)}{p(x)}$.
These classifiers are typically trained by maximizing the conditional logarithmic likelihood: $\max_{\theta} \mathbb{E}_{(x, y) \sim p(x, y)} \log p(x|y)$.

When using conditional diffusion models, $p(x|y)$ can be computed with the instantaneous change-of-variable formula \citep{Chen2018,Grathwohl2018} or approximated by ELBO \citep{sohl2015deep,ho2020denoising}. 
\citet{duvenaud2020your} advocate an energy-based model to improve calibration, robustness, and out-of-distribution detection. 
\citet{yang2022your} propose HybViT, which combines diffusion models and discriminative classifiers to predict $p(y|x)$. 
\citet{zimmermann2021score} introduce score-based generative classifiers but find them vulnerable to adversarial attacks. 
\citet{clark2024text} and \citet{li2023your} explore zero-shot generative classifiers using pretrained models like Stable Diffusion \citep{rombach2022high} to denoise images based on text labels.

In contrast, our approach fine-tunes a classifier on top of the generative model, making it a discriminative classifier that directly estimates $p(y|x)$. 
Discriminative classifiers are often preferred for classification tasks \citep{Vapnik1995,AndrewNIPS2001}. 
\citet{fetaya2019understanding} suggest a trade-off between likelihood-based modeling and classification accuracy, 
implying that improving generation may hinder classification for generative classifier.
However, our two-stage training scheme shows that enhancing the generative model in the pretraining stage can improve discriminative performance during the fine-tuning stage. 
This suggests a complementary relationship between likelihood-based training and classification accuracy, indicating that discriminative models can benefit from generative pretraining. 
Thus, we demonstrate a viable approach to leveraging generative models for discriminative tasks.

\paragraph{Representation Learning via Denoising Autoencoders}
Generative models have been widely explored for visual representation learning \citep{kingma2013auto,van2017neural,goodfellow2020generative}. 
Denoising autoencoders (DAEs) \citep{vincent2008extracting,vincent2010stacked,bengio2013generalized,geras2014scheduled} learn robust representations by reconstructing corrupted input data, which can be interpreted through a manifold learning perspective \citep{vincent2011}. 
Our method extends \citet{vincent2010stacked} by leveraging reversible flow-based generative models like diffusion models.

Diffusion models \citep{hu2025dynamicidzeroshotmultiidimage,sohl2015deep,ho2020denoising,song2021maximum}
%\citep{sohl2015deep,ho2020denoising,song2021maximum}
have shown promise in generation, yet their potential for feature extraction is underexplored. 
\citet{chen2020generative} pretrain a Transformer for autoregressive pixel prediction, demonstrating competitive results in representation learning. 
Their two-stage approach (pre-training a generative model followed by full fine-tuning) closely resembles ours.
Additionally, their downsampling of images to low resolution mimics the effect of adding noise, analogous to denoising. 
However, their model and autoregressive generation scheme are unsuitable for reverse generation and lack efficiency.
\citet{he2022masked} pretrain a masked autoencoder, effectively performing denoising by reconstructing incomplete data. 
\citet{baranchuk2021label} show diffusion models improve semantic segmentation, particularly with limited labeled data. 
\citet{sinha2021dc} propose Diffusion-Decoding with contrastive learning to enhance representation quality, while \citet{mittal2023diffusion} introduce a Diffusion-based Representation Learning (DRL) framework, leveraging denoising score matching.

% Recently, \citet{xiang2023denoising} investigate using denoising diffusion autoencoders for discriminative tasks via generative pre-training. 
% Our method differs by using the reverse generative process to extract features, offering a more general and model-agnostic approach.

Compared to previous methods, \textbf{our approach proposes to invert a pretrained continuous-time flow/diffusion generator: by running the model backward, it produces multi-level features that, after light fine-tuning, serve as an unsupervised representation extractor.}

%% file: sec/4_method.tex
\section{Method}

To address the issues discussed in Section~\ref{sec:related},
We propose an approach that follows the conventional two-stage training scheme commonly used in generative modeling applications.
In the first stage, we pretrain diffusion or flow models in an unsupervised manner.
In the second stage, we fine-tune the models for discriminative tasks by conducting inference in the reverse direction of the generative process.

\subsection{Pretrained Reversible Generation Framework}
We consider two continuous-time stochastic flow generative models: diffusion and flow models.  
Let $t$ denote time, where $t \in [1,0]$ corresponds to the generation, and $t \in [0,1]$ to its reverse. We denote the data and their transformations as $x_t$, with $x_0$ as the original and $x_1$ as the terminal state.

A diffusion model reverses a forward diffusion process described by $p(x_t \mid x_0) \sim \mathcal{N}(x_t \mid \alpha_t x_0, \sigma_t^2 I)$,
where $\alpha_t$ is a scaling factor and $\sigma_t^2$ denotes the variance.  
We adopt the Generalized VP-SDE, which uses a triangular function as coefficients to define the diffusion path, referred to as \textbf{PRG-GVP} with $\alpha_t = \cos\left(\frac{1}{2} \pi t\right)$ and $\sigma_t = \sin\left(\frac{1}{2} \pi t\right)$.
Due to the Fokker-Planck equation, an equivalent ODE can be derived that shares the same marginal distribution, known as the Probability Flow ODE \citep{song2020score}:
\begin{equation}\label{eq:reverse_diffusion_ode_velocity}
\small
    \frac{\mathrm{d}x_t}{\mathrm{d}t} = v(x_t) = f(t)x_t - \frac{1}{2} g^2(t) \nabla_{x_t} \log p(x_t).
\end{equation}

A flow model deterministically transforms $x_0$ to $x_1$ via a parameterized velocity function $v(x_t)$.  
Inspired by \citep{tong2024improving}, we adopt the flow path formulation, referred to as \textbf{PRG-ICFM}, with $p(x_t | x_0, x_1) = \mathcal{N}(x_t | t x_1 + (1 - t) x_0)$, 
and velocity function: $v(x_t | x_0, x_1) = x_1 - x_0$.

We also employ the Optimal Transport Conditional Flow Matching (OTCFM) model \citep{tong2024improving}. Specifically, we jointly sample $(x_0, x_1)$ from the optimal transport plan $\pi$ to construct the 2-Wasserstein target distribution. The corresponding velocity field $v(x_t \mid x_0, x_1)$ is then used to approximate dynamic optimal transport, referred to as \textbf{PRG-OTCFM}.
In practice, we parameterize the velocity model $v(x_t)$ as $v_{\theta}$ using a network to model the above variants of PRG.
\begin{equation}
    \label{eq:matching_loss}
    \small
    \begin{aligned}
    \mathcal{L}_{\text{FM}} &= \frac{1}{2} \mathbb{E}_{p(x_t)}\left[\lambda_{\mathrm{FM}}(t)\|v_\theta(x_t) - v(x_t)\|^2\right] \mathrm{d}t
    \end{aligned}
\end{equation}

After completing pre-training via \cref{eq:matching_loss}, and given the invertibility of \cref{eq:reverse_diffusion_ode_velocity}, we obtain $x_1$ for a data point $x_0$ through the same reversal. We denote $x_t = F_{\theta}(x_0)$ as the extracted features for downstream discriminative tasks, such as classification or regression.

This paper focuses on supervised learning for downstream tasks while simultaneously improving the lower bound of mutual information. Suppose we have a labeled dataset $\mathcal{D}_{\text{finetune}} = \{(x_i, y_i)\}_{i=1}^{N}$, where $x_i$ is a data point and $y_i$ is its label, potentially differing from the pre-training dataset $\mathcal{D}_{\text{pretrain}}$. Thus, during the fine-tuning stage, we introduce a classifier $p_{\phi}(y \mid z)$, parameterized by $\phi$, minimizing the cross-entropy loss and flow-matching loss:
\begin{equation}\label{eq:cross_entropy_loss}
\small
    \mathcal{L}_{\text{total}} = -\sum_{i=1}^{N} \log p_{\phi}(y_i \mid F_{\theta}(x_i)) + \beta\mathcal{L}_{\text{FM}}(x).
\end{equation}
\paragraph{Classifier Design}
We found that the features $x_t$ are well-structured, allowing a simple two-layer MLP with tanh activations to perform well. Notably, enlarging the classifier yielded minimal gains, suggesting that the reversed generative process already captures highly informative features.
\paragraph{Diffusion Step Design}
To reduce computational costs, the representations used for inference do not necessarily need to traverse the entire process. 
Instead, our framework allows fine-tuning and inference to start from any selected point along the trajectory. Moreover, more complex downstream tasks typically require greater training/inference steps. Further details are provided in \cref{sec:my432,sec:Ablation Studies appendix}.
\subsection{Pretrained Reversible Generation Mechanism}
\label{sec:theroy}
\paragraph{pre-training}
In the traditional autoencoder paradigm, representation learning aims to learn a network that maps the input data $X$ into a useful representation $Z$ for downstream tasks. 
We parameterize the encoder $p_{\theta}(z|x)$ with parameters $\theta$. 
A good representation should retain sufficient information about the input $X$. 
In information-theoretic terms \citep{linsker1988application,vincent2010stacked}, this corresponds to maximizing the mutual information $\mathcal{I}(X,Z)$ between the input random variable $X$ and its representation $Z$.
Following the reasoning of \citet{vincent2010stacked}, we show that pre-training diffusion or flow models maximizes the mutual information.
Specifically, we aim to find the optimal $\theta^*$ that maximizes $\mathcal{I}(X,Z)$:
\begin{equation}\label{eq:maximize_mutual_information}
    \begin{aligned}
    \small
        \theta^* = \arg\max_{\theta} \mathcal{I}(X,Z) = \arg\max_{\theta} [\mathcal{H}(X) - \mathcal{H}(X|Z)] \\
        = \arg\max_{\theta} [-\mathcal{H}(X|Z)] = \arg\max_{\theta} \mathbb{E}_{p(z,x)} [\log p(x|z)].
    \end{aligned}
\end{equation}

% Since $\mathcal{H}(X)$ is constant with respect to $\theta$, maximizing $\mathcal{I}(X,Z)$ is equivalent to minimizing the conditional entropy $\mathcal{H}(X|Z)$.
% The last equality follows from the definition of conditional entropy.

Suppose we have a decoder $\theta'$, which approximates $p_{\theta'}(x|z)$ to recover data $x$ from the latent variable $z$.
By the non-negativity of the KL divergence, we have:
\begin{equation}\label{eq:KL}
\small
    \mathbb{D}_{\mathrm{KL}} [p(x|z)||p_{\theta'}(x|z)] \ge 0.
\end{equation}
\begin{equation}\label{eq:lower_bound_mutual_information}
\small
    \mathbb{E}_{p(z,x)} [\log p(x|z)] \ge \mathbb{E}_{p(z,x)} [\log p_{\theta'}(x|z)].
\end{equation}
Thus, the right term serves as a lower bound on $\mathcal{I}(X,Z)$.

In our setting, we consider deterministic mappings for both the encoder and decoder, which are invertible.
Let us denote the mapping $z = F_{\theta}(x)$, which implies $p_{\theta}(z|x) = \delta(z - F_{\theta}(x))$,
which is Dirac delta function. 
% Similarly, the decoder mapping is $x = G_{\theta'}(z)$, with $p_{\theta'}(x|z) = \delta(x - G_{\theta'}(z))$.
Under these assumptions, we can rewrite the RHS of \cref{eq:lower_bound_mutual_information} as:

\begin{equation}\label{eq:reconstruction_error}
\small
    \begin{aligned}
        \mathbb{E}_{p(z,x)}{[\log{p_{\theta'}(x|z)}]}=\mathbb{E}_{p(x)}{[\log{p_{\theta'}(x|z=F_{\theta}(x))}]}\\
        =-\mathbb{D}_{\mathrm{KL}}(p(x)||p_{\theta'}(x|z=F_{\theta}(x)))-\mathcal{H}(X)\\
    \end{aligned}
\end{equation}

where $p(x)$ is an unknown data distribution that we can sample from. Eq.\ref{eq:reconstruction_error} is for maximizing the log likelihood of reconstructing data $X$,
or equivalently, minimizing KL divergence between the data and the generated distribution.

In continuous-time stochastic flow generative models, the inference and generation process are reversible via a same model, which means the decoder parameter $\theta'$ and encoder parameter $\theta$ are shared parameters.
However, we still use these two notation in the later part of this paper for distinguishing the inference and generation process through ODE solver with $\theta$ and $\theta'$ respectively.
For preventing any confusion, we denote $x_0$ being data variable $x$ amd $x_1$ being the inferred latent variable $z$.
In practice, training generative models using score matching or flow matching objectives does not directly maximize the likelihood of reconstructing $x_0$ from $x_1$.
However, as shown in previous works~\citep{song2021maximum, lu2022maximum, zheng2023improved}, these methods provide a lower bound on the likelihood.
Specifically, they establish that:
\begin{equation}\label{eq:maximize_likelihood_and_score_matching}
\small
    \begin{aligned}
        & \mathbb{D}_{\mathrm{KL}}(p(x_0)||p_{\theta'}(x_0))=\mathbb{D}_{\mathrm{KL}}(p(x_1)||p_{\theta'}(x_1))+\mathcal{L}_{\text{ODE}}\\
        & \le \mathbb{D}_{\mathrm{KL}}(p(x_1)||p_{\theta'}(x_1))+\sqrt{\mathcal{L}_{\text{SM}}}\sqrt{\mathcal{L}_{\text{Fisher}}}.
    \end{aligned}
\end{equation}
The term $\mathbb{D}_{\mathrm{KL}}(p(x_1)||p_{\theta'}(x_1))\approx 0$ because both $p(x_1)$ and $p_{\theta'}(x_1)$ are approximately the same Gaussian distribution. 
The losses $\mathcal{L}_{\text{ODE}}$, $\mathcal{L}_{\text{SM}}$ and $\mathcal{L}_{\text{Fisher}}$ are defined as:

\begin{equation}\label{eq:notation}
\small
    \begin{aligned}
        \mathcal{L}_{\text{ODE}} &=\frac{1}{2}\mathbb{E}_{p(x_t)}[g^2(t)(s_{\theta}(x)-\nabla\log p(x_t))^T\\
        &(\nabla\log p_{\theta'}(x_t)-\nabla\log p(x_t))]\mathrm{d}t\\
        \mathcal{L}_{\text{SM}} &=\frac{1}{2}\mathbb{E}_{p(x_t)}\left[g^2(t)\|s_{\theta}(x)-\nabla\log p(x_t)\|^2\right]\mathrm{d}t\\
        \mathcal{L}_{\text{Fisher}} &=\frac{1}{2}\mathbb{E}_{p(x_t)}\left[g^2(t)\|\nabla\log p_{\theta'}(x_t)-\nabla\log p(x_t)\|^2\right]\mathrm{d}t,\\
    \end{aligned}
\end{equation}
where $g(t)$ is a special weighting function, $s_{\theta}(x_t)$ is the score function parameterized by $\theta$, and $p(x_t)$ is the data distribution at time $t$ along forward diffusion process.

From \cref{eq:maximize_likelihood_and_score_matching}, we observe that using score matching objectives during training can help increase the data likelihood. 
Flow matching differs from score matching by employing a slightly different weighting factor in the loss function, so it can also be used to increase the likelihood, as shown in \cref{eq:matching_loss}.
Therefore, pre-training diffusion or flow models via score matching or flow matching can be viewed as maximizing the mutual information between the data and its representation encoded through the reverse generative process.
This relationship between the mutual information and the training loss is guaranteed by the \cref{eq:maximize_likelihood_and_score_matching}.
This process transforms the data distribution into a Gaussian distribution while preserving as much information as possible.
\paragraph{Fine-tuning by Reversing Generation}
Generative features may not be optimal for discriminative tasks, as reconstruction requires more detail than comprehension. Hence, fine-tuning is necessary to adapt the model for downstream tasks.
\textbf{A key question is whether the flow model should also be fine-tuned alongside the classifier.} Due to the flow model's large capacity, reusing the pretrained model seems preferable. Initially, we explored freezing the model while fine-tuning only the classifier. However, this approach underperformed, indicating that fine-tuning both the flow model and the classifier is necessary. Otherwise, a large and complex classifier would be required, reducing efficiency.
To fine-tune the entire model, we leverage neural ODEs \citep{Chen2018} for efficient gradient computation, optimizing both $\theta$ and $\phi$ via backpropagation.
We observed that distinct discriminative tasks correspond to different parts of generative features~\cref{sec:Reverse Generation Process}. Thus, fine-tuning can selectively prune or enhance relevant features as needed.
In addition, we assess the robustness and generalization of these features against out-of-distribution samples through further experiments.

\begin{figure*}[!h]
  \centering
  \begin{minipage}[t]{0.32\linewidth}
    \centering
    \small
    \includegraphics[width=\linewidth]{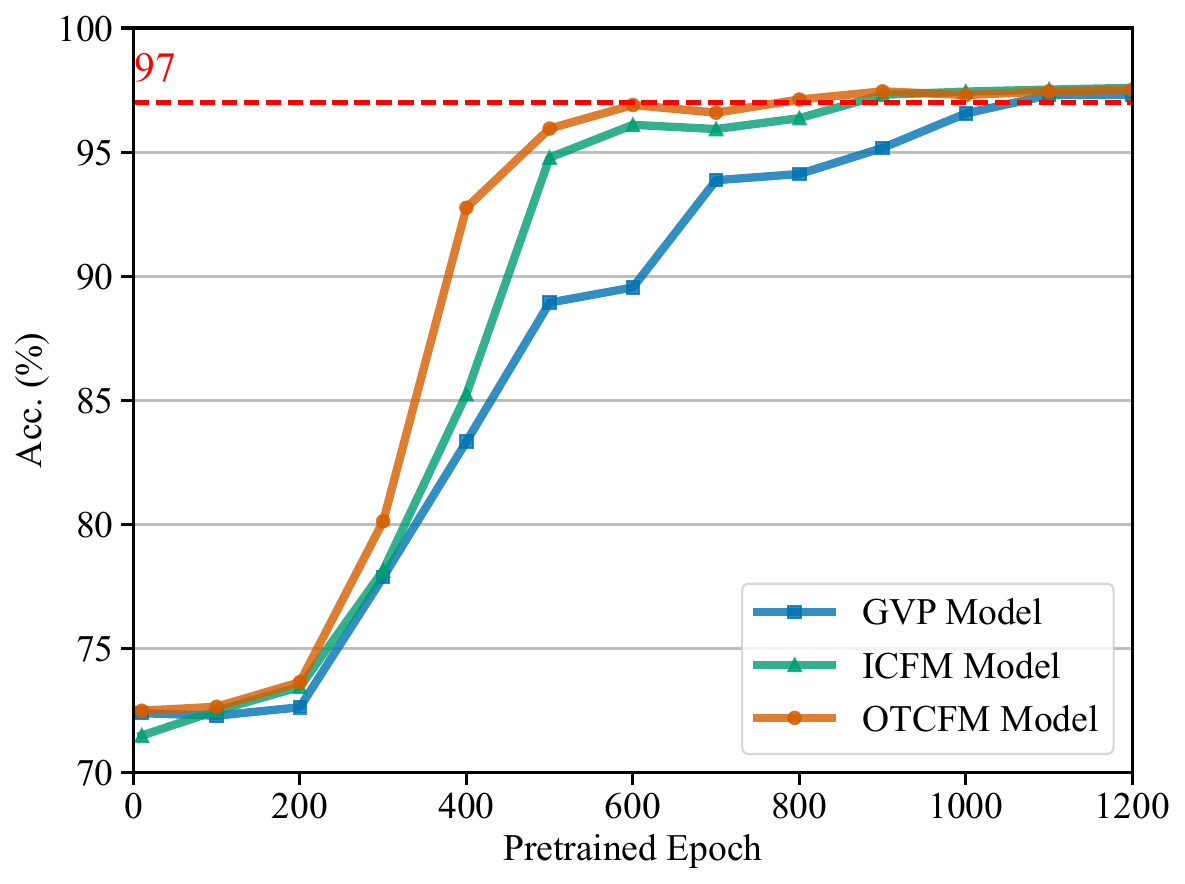}
    \caption{CIFAR-10 Accuracy (y-axis) after fine-tuning checkpoints from different pre-training stages (x-axis).} 
    \label{fig:cifar-acc}
  \end{minipage}
  \hfill
  \begin{minipage}[t]{0.32\linewidth}
    \centering
    \small
    \includegraphics[width=\linewidth]{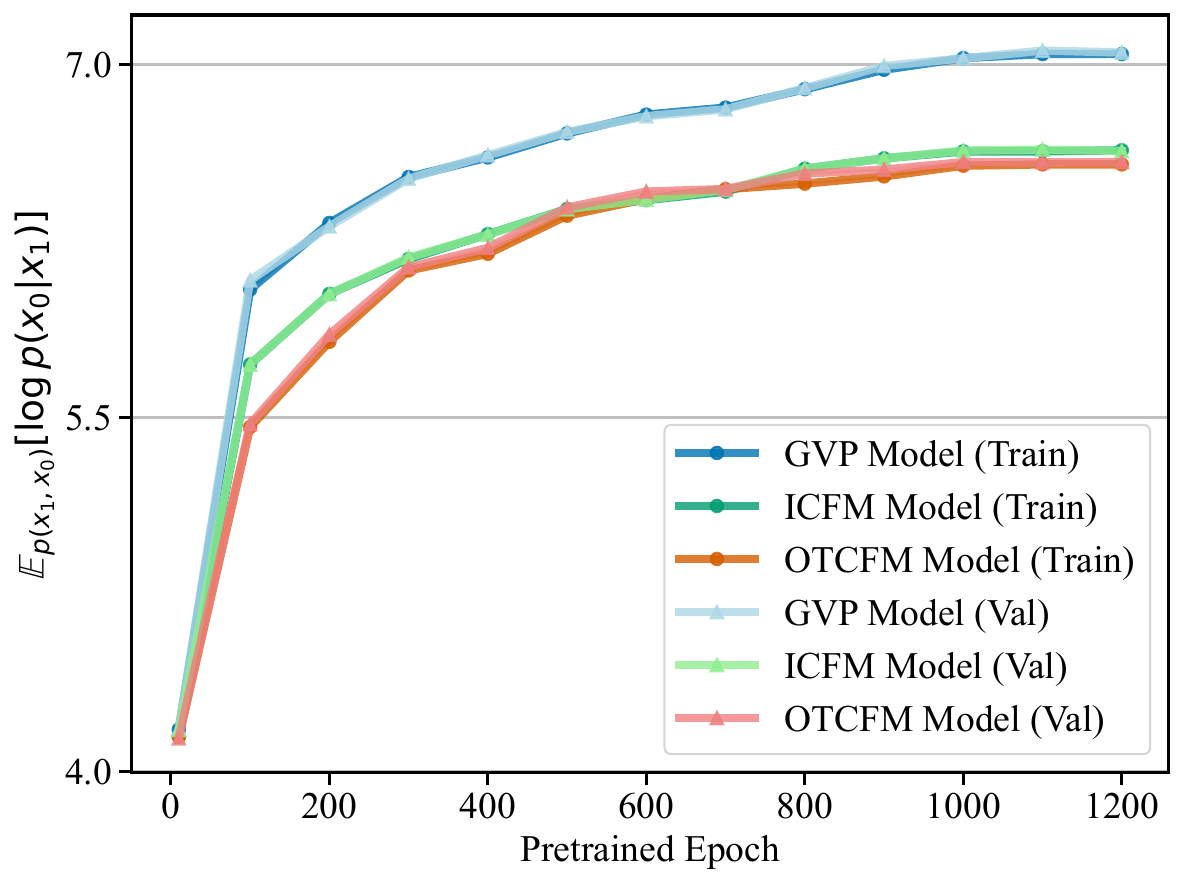}
    \caption{Evaluation of the indicator (y-axis) $\mathbb{E}_{p(x_1,x_0)} [\log p(x_0|x_1)]$ from different pre-training stages (x-axis).}
    \label{fig:cifar-logp}
  \end{minipage}
  \hfill
  \begin{minipage}[t]{0.32\linewidth}
    \centering
    \small
    \includegraphics[width=\linewidth]{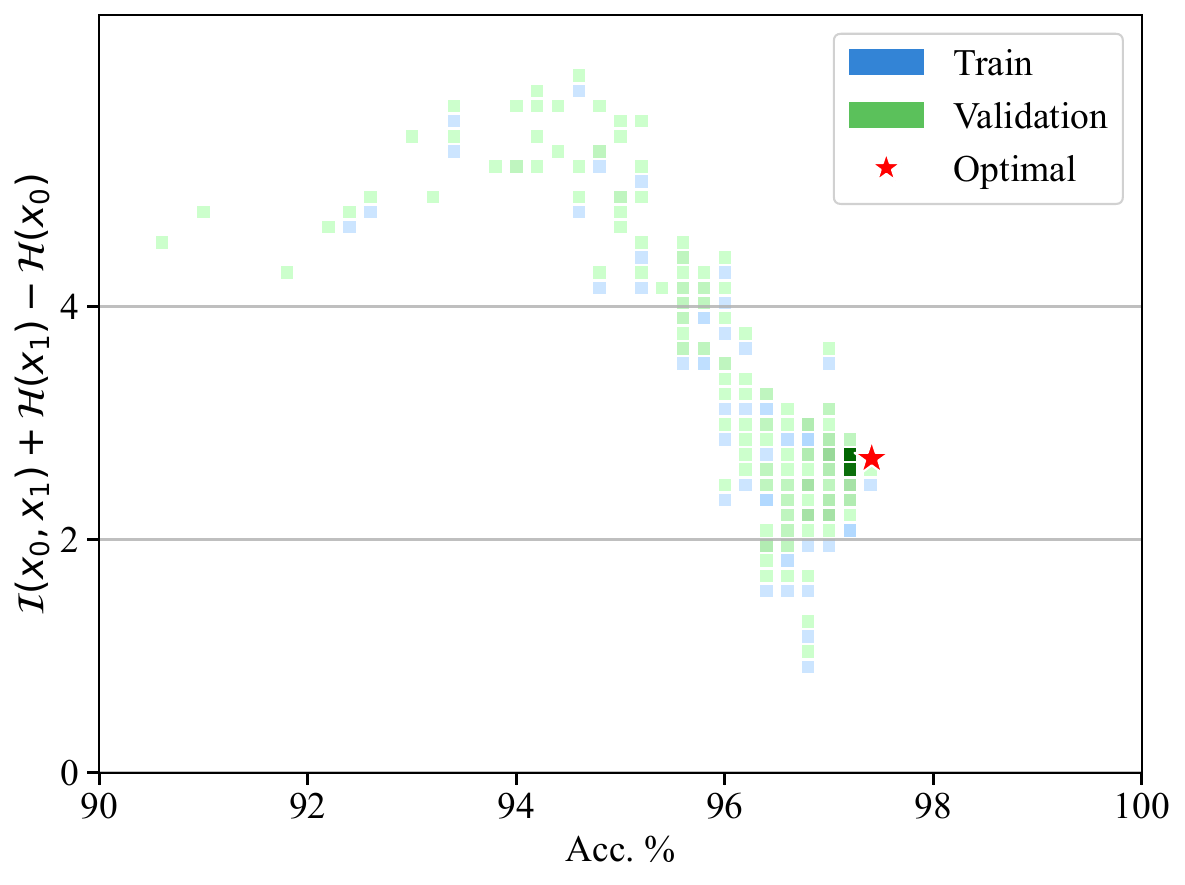}
   \caption{The relationship between y-axis $\mathcal{I}(X_0, X_1) + \mathcal{H}(X_1) - \mathcal{H}(X_0)$ and accuracy (x-axis) during fine-tuning.}
    \label{fig:logp-acc}
  \end{minipage}
\end{figure*}
\subsection{Advantages}
\paragraph{Model-Agnostic Flexibility}
Our approach is agnostic to the choice of network architecture for the generative model: it can be a U-Net, Transformer, or any other model. The latent variable $Z$ remains stationary with respect to the data $X$ when using an ODE solver due to the static nature of continuous-time stochastic flow models~\citep{song2020score}. 
This allows different architectures to encode the data while yielding the same latent variable $Z$. 
In contrast, methods like DDAE~\citep{xiang2023denoising} are model-dependent, with the latent variable $Z$ tied to specific network architectures and derived from intermediate activations, making them less flexible and non-stationary.

% \vspace{-10pt}

\paragraph{Infinite-Layer Expressiveness}
PRG leverages the infinite-layer structure of continuous-time stochastic flow models \citep{Chen2018}, providing high expressiveness with relatively small parameter sizes. 
This structure underpins the success of modern generative models like SD3 \citep{rombach2022high} and DALL·E 3 \citep{betker2023improving}. 
By using a reversible process, our method allows for the simultaneous training of discriminative and generative models, offering excellent performance with enhanced capacity.

% \vspace{-10pt}

\paragraph{Robustness and Generalizability}
Features extracted by reversing the generative process remain stable along the reverse trajectory, allowing feature use from any point \( t \in [0, 1] \) for downstream tasks, not just \( x_1 \). These features are robust to various discretization schemes (e.g., Euler, Runge-Kutta) and time steps (\( 10^3 \) to \( 10^1 \)). Our method generalizes well across datasets, enabling fine-tuning of pretrained models for new tasks. Notably, it applies to all diffusion and flow models and has been successfully transferred to community-developed text-to-image models in \cref{sec:trans}.

A more detailed analysis of why PRG is effective can be found in \cref{sec:myadvantage}.

%% file: sec/5_exp.tex
\section{Experiments}
%-------------------------------------------------------------------------
\subsection{Settings}
For generative model selection, we adopt three diffusion and flow-based models: GVP, ICFM, and OTCFM.
For pre-training, we followed the protocol from \cite{nichol2021improved}, using the Adam optimizer~\cite{kingma2014adam} with a fixed learning rate of $1 \times 10^{-4}$ for 1,200 epochs on the respective training sets. 
For downstream image classification, we adopted the configuration from \cite{liu2021swin}, using AdamW for 200 epochs with a cosine decay learning rate schedule and a 5-epoch linear warm-up. The training used a batch size of 128 and an initial learning rate of 0.001.
We followed DDAE~\cite{xiang2023denoising}'s setting (64 resolution, simple data augmentation, no mixup~\cite{zhang2017mixup}, no cutmix~\cite{yun2019cutmix}) for fair comparison, yet our method achieves SOTA diffusion classifier performance. 
Further details, \textbf{including hyperparameters and efficiency statistics for training and inference}, are provided in \cref{sec:Training Process Details,sec:Evaluation of Training Efficiency,sec:evaluation_inference_efficiency} .
\subsection{baselines}
We select state-of-the-art (SOTA) representation models as baselines to evaluate PRG on CIFAR-10~\cite{Krizhevsky2009LearningML}, Tiny-ImageNet~\cite{Le2015TinyIV}, and ImageNet~\cite{Russakovsky2014ImageNetLS}. Baselines include classical discriminative models like WideResNet-$28$-$10$~\cite{zagoruyko2016wide} and ResNeXt-$29$-$16\times64$d~\cite{xie2017aggregated}, as well as diffusion-based generative models such as GLOW~\cite{fetaya2019understanding}, Energy Model~\cite{duvenaud2020your}, SBGC~\cite{zimmermann2021score}, HybViT~\cite{yang2022your}, and DDAE~\cite{xiang2023denoising}.
For out-of-distribution (OOD) evaluation, we compare PRG with PGD~\cite{madry2017towards}, PLAT~\cite{wang2020enresnet}, AugMix~\cite{hendrycks2019augmix}, AutoAug~\cite{cubuk2019autoaugment}, SBGC, and PDE+~\cite{yuan2024pde} on CIFAR-10-C and Tiny-ImageNet-C.
All results are from their original papers.

\subsection{Verification Analysis}
\label{sec:Verification Analysis}
\subsubsection{Better Pre-training Lead to Better Fine-tuning?}
To explore whether better pre-training improves fine-tuning, we compared the classification accuracies of models finetuned from different pre-training epochs on CIFAR-10 (\cref{fig:cifar-acc}). 
We also computed the mutual information during pre-training, which increases monotonically, as shown in \cref{fig:cifar-logp}. During pre-training, we have
\[
\small
\mathbb{E}_{p(x_1,x_0)}\bigl[\log p(x_0 \mid x_1)\bigr]
= \mathcal{I}(X_0, X_1) \;-\; \mathcal{H}(X_0),
\]
where \(\mathcal{H}(X_0)\) is the (constant) entropy of the data distribution and \(\mathcal{H}(X_1)\) is the (constant) entropy of the Gaussian distribution. Hence, \(\mathbb{E}_{p(x_1,x_0)}[\log p(x_0 \mid x_1)]\) faithfully reflects the mutual information \(\mathcal{I}(X_0, X_1)\) during pre-training.
In contrast, during fine-tuning, \(\mathbb{E}_{p(x_1,x_0)}[\log p(x_0 \mid x_1)]\) no longer represents \(\mathcal{I}(X_0, X_1)\), because \(\mathcal{H}(X_1)\) is no longer the entropy of a Gaussian distribution. Nonetheless, the quantity
$\mathcal{I}(X_0, X_1) + \mathcal{H}(X_1) \;-\; \mathcal{H}(X_0)$
still captures the change in \(\mathcal{I}(X_0, X_1)\) to a large extent.
We compute log-probability density $\log p(x_0|x_1)$ using the adjoint method or Neural ODEs~\citep{Chen2018} for computational feasibility:  
\begin{equation}\label{eq:neural_ode}
\small
    \begin{aligned}
    \begin{bmatrix} x_0 - x_1\\ \log{p(x_0)}-\log{p(x_1)} \end{bmatrix} = \int_{t=1}^{t=0} \begin{bmatrix} v_{\theta}(x_t, t) \\ - \text{Tr} \left( \frac{\partial v_{\theta}}{\partial x_t} \right) \end{bmatrix} \mathrm{d}t
    \end{aligned}
\end{equation}
For efficient computation of $\log p(x_0|x_1)$, we employ the Hutchinson trace estimator \cite{Hutchinson1989,Grathwohl2018}:  
\begin{equation}\label{eq:hutchinson}
\small
    \begin{aligned}
        \text{Tr} \left( \frac{\partial v_{\theta}}{\partial x_t} \right) = E_{p(\epsilon)}[\epsilon^T \frac{\partial v_{\theta}}{\partial x_t} \epsilon]
    \end{aligned}
\end{equation}
where $\epsilon$ is a standard Gaussian random vector.

\begin{figure}
  \centering
  \small
    \includegraphics[width=\linewidth]{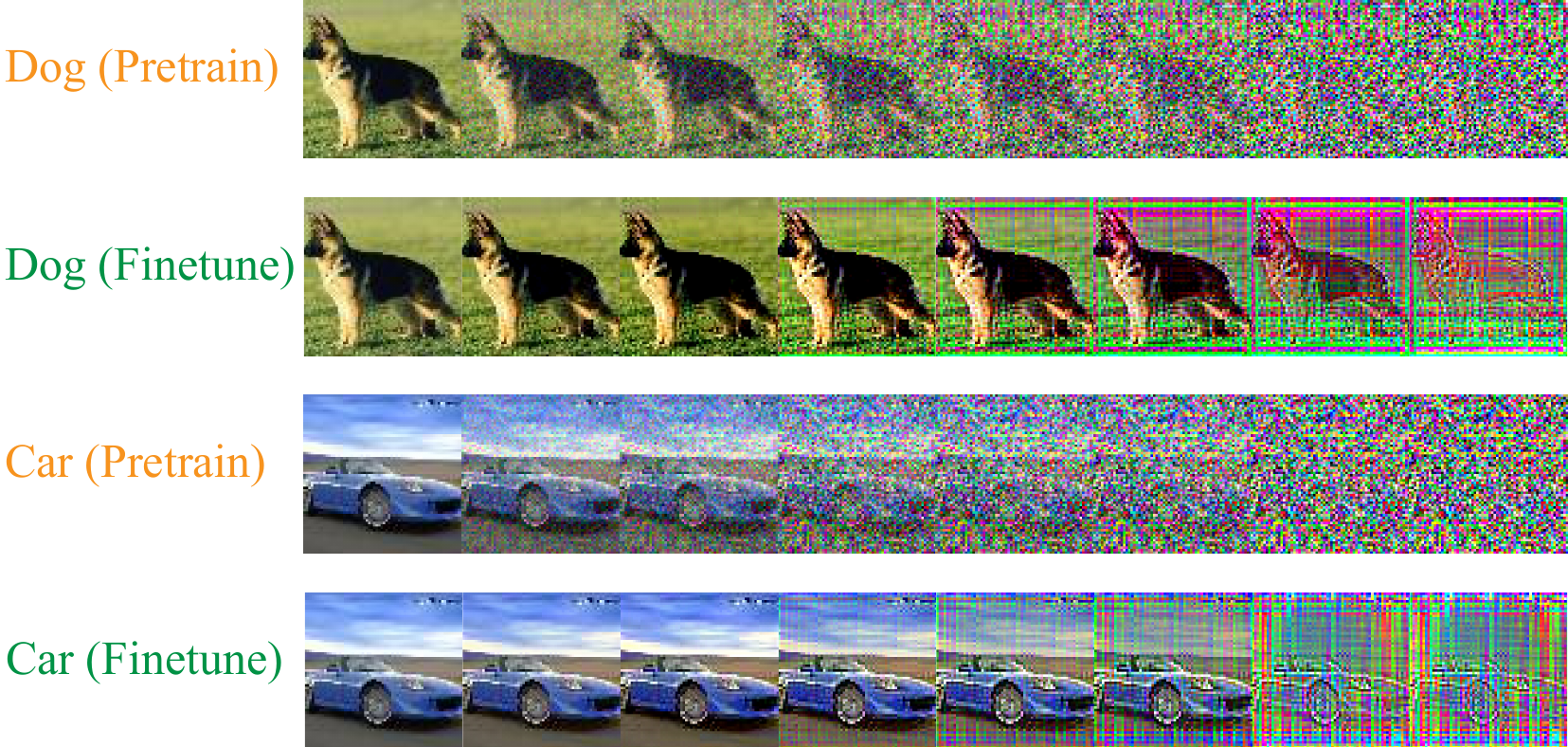}
    \caption{Reverse generation from \(x_0\) to \(x_1\) with full pre-training and fine-tuning. }
    \label{fig:epoch-show}
    %explan the picture
\end{figure}

Moreover, we analyzed the relationship between accuracy and mutual information during fine-tuning, as shown in \cref{fig:logp-acc}.
Models without pre-training achieved approximately $73.5\%$ accuracy. In contrast, models with more pre-training exhibit higher mutual information and classification accuracy, suggesting that stronger generative capability enhances fine-tuning performance.
During fine-tuning, we observed a decline in $\mathcal{I}(x_0, x_1) + \mathcal{H}(x_1) - \mathcal{H}(x_0)$ for both the training and validation sets, suggesting that the model filters out unnecessary features to improve downstream classification,
which can also be observed in \cref{fig:epoch-show} as some visual representations are kept while others are discarded.
In summary, \textbf{sufficient pre-training is crucial for optimal performance in inverse generative classification.}

\subsubsection{What is a Reasonable Fine-tuning Strategy?}
\label{sec:my432}
\begin{figure}
  \centering
  \small
  \includegraphics[width=0.9\linewidth]{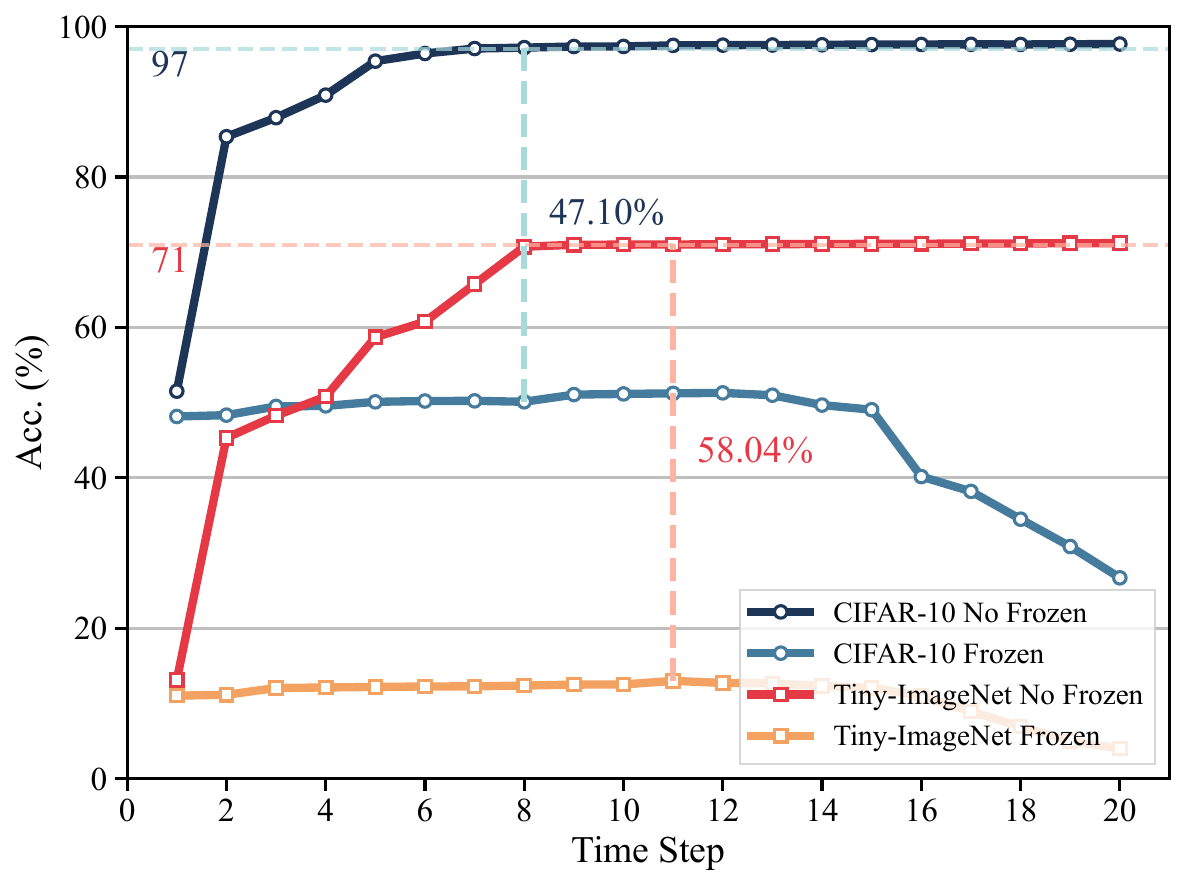}
  \caption{Accuracy on CIFAR-10 and Tiny-ImageNet when fine-tuned from different time steps along the generation trajectory.}
  \label{fig:combine_acc}
\end{figure}

We first examine whether a frozen flow model provides meaningful features. 
For this, we use features $x_t$ along the sampling trajectory from $x_0$ to $x_1$ under two conditions: (1) freezing the generative model's parameters (light lines) and (2) end-to-end training of both the generative model and the classification head (dark lines). 
As shown in \cref{fig:combine_acc}, the performance gap is substantial—47.10\% on CIFAR-10 and 58.04\% on Tiny-ImageNet—demonstrating the \textbf{critical importance of updating the generative model's parameters during fine-tuning}.

Additionally, on CIFAR-10, initiating fine-tuning from later stages of the sampling trajectory ($x_{1/4}$ to $x_1$) results in progressively better performance. 
On TinyImageNet, optimal results are achieved when fine-tuning starts from $x_{1/2}$ to $x_1$. 
This difference likely stems from the complexity of the datasets: \textbf{more complex datasets require stronger feature extraction, necessitating longer trajectory lengths for effective fine-tuning.} For an in-depth analysis, see \cref{sec:without pre-training}; complementary experiments on optimal fine-tuning strategies are provided in \cref{sec:Ablation Studies appendix}.

\subsubsection{Is the Model a Continuous Feature Extractor?}
Residual networks, such as ResNet-50~\cite{he2016deep}, achieve feature extraction by constructing discrete-time methods through a series of composite transformations: $\mathbf{h}_{t+1} = \mathbf{h}_t + f(\mathbf{h}_t, \theta_t)$. 
% \begin{equation}
%   \begin{aligned}
%     \mathbf{h}_{t+1} &= \mathbf{h}_t + f(\mathbf{h}_t, \theta_t)
%   \end{aligned}
%   \label{eq:disrect method}
% \end{equation}
In contrast, we construct the feature extractor using the ODE specified by the neural network, namely $\frac{d\mathbf{h}(t)}{dt} = f(\mathbf{h}(t), t, \theta)$.
% \begin{equation}
%   \begin{aligned}
%     \frac{d\mathbf{h}(t)}{dt} &= f(\mathbf{h}(t), t, \theta)
%   \end{aligned}
%   \label{eq:ode}
% \end{equation}
\cref{tab:sampling_length,tab:model_distance}  demonstrate that although we trained using a discrete sample length of $t_{\text{span}} = 20$, we can evaluate the model using any reasonable inference steps. Furthermore, feature extraction is not limited to a specific training point (such as the midpoint); extracting features within its vicinity (within 20\%) still achieves high performance on downstream tasks.
\textbf{These findings validate our method can serve as a continuous feature extractor.}

\begin{table}[h]
  \centering
  \small
  \begin{tabular}{@{}lccccc@{}}
    \toprule
    \textbf{Inference Steps} & \textbf{20} & \textbf{100} & \textbf{500} & \textbf{800} & \textbf{1000}\\
    \midrule
    PRG-GVP Acc. (\%) & 97.25 & 97.28 & 97.30 & 97.31 & 97.31 \\
    PRG-ICFM Acc. (\%) & 97.32 & 97.33& 97.34& 97.34  & 97.34 \\
    PRG-OTCFM Acc. (\%) & 97.42 & 97.43& 97.43& 97.44  & 97.44 \\
    \bottomrule
  \end{tabular}
    \caption{The checkpoint, fine-tuned at the midpoint of a trajectory with $t_{\text{span}}=20$, is evaluated on CIFAR-10 using the midpoints of trajectories with varying inference steps (from 20 to 1000).}
  \label{tab:sampling_length}
\end{table}

\begin{table}[h]
  \centering
  \small
  \begin{tabular}{@{}lccccc@{}}
    \toprule
    \textbf{Offset from Midpoint \%} & \textbf{-20} & \textbf{-10} & \textbf{10} & \textbf{20} \\
    \midrule
    PRG-GVP Acc. (\%) & $97.02$ & $97.17$  & $97.12$ & $97.08$ \\
    PRG-ICFM Acc. (\%) & $97.23$ & $97.30$  & $97.20$ & $97.10$ \\
    PRG-OTCFM Acc. (\%) & $97.31$ & $97.32$ & $97.26$ & $97.13$ \\
    \bottomrule
  \end{tabular}
  \caption{Accuracy at the midpoints of trajectories with varying offsets for $t_{\text{span}} = 1000$ (use the same checkpoint from \cref{tab:sampling_length}).}
  \label{tab:model_distance}
\end{table}

%----------------------------------------------------------------------------

\begin{table}[h]
  \centering
  \small
  \begin{tabular}{@{}lcc@{}}
    \toprule
    \textbf{Method} & \textbf{Param. (M)} & \textbf{Acc. (\%)} \\
    \midrule
    \multicolumn{3}{c}{\textbf{Discriminative  methods}} \\
    WideResNet-$28$-$10$ \cite{zagoruyko2016wide} & $36$ & $96.3$ \\
    ResNeXt-$29$-$16*64$d \cite{xie2017aggregated} & $68$ & $96.4$ \\
    \midrule
    \multicolumn{3}{c}{\textbf{Generative methods}} \\
    GLOW \cite{fetaya2019understanding} & N/A & $84.0$ \\
    Energy model \cite{duvenaud2020your} & N/A & $92.9$ \\
    SBGC \cite{zimmermann2021score} & N/A & $95.0$ \\
    HybViT \cite{yang2022your} & $43$ & $96.0$ \\
    DDAE \cite{xiang2023denoising} & $36$ & $97.2$ \\
    \hdashline
    \multicolumn{3}{c}{\textbf{Our methods}} \\
    PRG-GVP-onlyPretrain & $42$ & $54.10$ \\
    PRG-GVP-S & $42$ & $97.35$ \\
    PRG-ICFM-S & $42$ & $97.59$ \\
    PRG-OTCFM-S & $42$ & $97.65$ \\
    \bottomrule
  \end{tabular}
  \caption{Comparison on the CIFAR-10 dataset with various algorithms. All results are reported from their original papers.}
  \label{tab:comparison on CIFAR-10}
\end{table}
\begin{table}[h]
  \centering
  \small
  \begin{tabular}{@{}lcc@{}}
    \toprule
    \textbf{Method} & \textbf{Param. (M)} & \textbf{Acc. (\%)} \\
    \midrule
    \multicolumn{3}{c}{\textbf{Discriminative  methods}} \\
    WideResNet-$28$-$10$ \cite{zagoruyko2016wide} & $36$ & $69.3$ \\
    
    \midrule
    \multicolumn{3}{c}{\textbf{Generative methods}} \\
    HybViT \cite{yang2022your} & $43$ & $56.7$ \\
    DDAE \cite{xiang2023denoising} & $40$ & $69.4$ \\
    \hdashline
    \multicolumn{3}{c}{\textbf{Our methods}} \\
    PRG-GVP-onlyPretrain & $42$ & $15.34$ \\
    PRG-GVP-S & $42$ & $70.98$ \\
    PRG-ICFM-S & $42$ & $71.12$ \\
    PRG-OTCFM-S & $42$ & $71.33$ \\
    \bottomrule
  \end{tabular}
  \caption{Comparison on the Tiny-ImageNet dataset with various algorithms. All results are reported from their original papers.}
  \label{tab:comparison on Tiny-ImageNet}
\end{table}

\begin{table}[h]
  \centering
  \small
  \begin{tabular}{@{}lcc@{}}
    \toprule
    \textbf{Method} & \textbf{Param. (M)} & \textbf{Acc. (\%)} \\
    \midrule
    \multicolumn{3}{c}{\textbf{Discriminative  methods}} \\
    ViT-L/$16$ ($384^2$) \cite{dosovitskiy2020image} &$307$ &$76.5$\\
    ResNet-$152$ ($224^2$) \cite{he2016deep} & $60$ & $77.8$ \\
    Swin-B ($224^2$) \cite{liu2021swin} & $88$& $83.5$\\
    \midrule
    \multicolumn{3}{c}{\textbf{Generative methods}} \\
    HybViT ($32^2$) \cite{yang2022your} & $43$ & $53.5$ \\
    DMSZC-DiTXL2 ($256^2$) \cite{li2023your}  & $338$ & $77.5$ \\
    iGPT-L ($48^2$) \cite{chen2020generative} & $1362$ & $72.6$ \\
    \hdashline
    \multicolumn{3}{c}{\textbf{Our methods}} \\
    PRG-GVP-onlyPretrain ($64^2$) & $122$ & $20.18$ \\
    PRG-GVP-XL ($64^2$) & $122$ & $77.84$ \\
    PRG-ICFM-XL ($64^2$) & $122$ & $78.12$ \\
    PRG-OTCFM-XL  ($64^2$) & $122$ & $78.13$ \\
    \bottomrule
  \end{tabular}
  \caption{Comparison on the ImageNet dataset with various algorithms. All results are reported from their original papers, the values in parentheses indicate the input image size.}
  \label{tab:model_comparison on ImageNet}
\end{table}

\subsection{Main Results}
\subsubsection{Performance on Image Classification}
\label{sec:unet}
To comprehensively evaluate our method, we conducted experiments on three image classification datasets,
as detailed in \cref{tab:comparison on CIFAR-10,tab:comparison on Tiny-ImageNet,tab:model_comparison on ImageNet}.
Our methods achieved accuracies of 97.59\%, 71.12\%, and 78.1\% on CIFAR-10, Tiny-ImageNet, and ImageNet,
respectively, surpassing all existing generative approaches, as well as supervised methods like WideResNet-28\cite{zagoruyko2016wide}.
However, these results still trail behind recent transformer-based supervised learning architectures like SwinTransformer~\cite{liu2021swin}.
We hypothesize that integrating transformer architectures and leveraging VAEs to handle high input resolutions could further enhance the performance.
This avenue will be explored in future work.

\subsubsection{Out-of-Distribution Robustness}
To address the degradation in out-of-distribution (OOD) due to common image corruptions or adversarial perturbations \cite{zimmermann2021score},
data augmentations and adversarial training are typically employed.
However, recent studies \cite{zimmermann2021score, chendiffusion, li2023your} have indicated that generative-based classifiers,
without requiring additional data, often exhibit superior OOD robustness.
As illustrated in \cref{tab:oodcifar} and \cref{sec:Details of Out-of-Distribution Experiments}, our method showcases remarkable robustness on these two datasets with only simple data augmentation.
\begin{table}
  \centering
  \small
  \begin{tabular}{@{}lcccc@{}}
    \toprule
    \textbf{Model} & \textbf{CIFAR-10} & \multicolumn{3}{c}{\textbf{CIFAR-10-C}} \\
    \cmidrule(lr){3-5}
     & \textbf{Clean} & \textbf{Corr Severity All} & \textbf{w/ Noise}  \\
    \midrule
    \multicolumn{4}{c}{\textbf{Adversarial Training}} \\
    PGD *\cite{madry2017towards} & $93.91$ & $83.08$ \scriptsize{$\downarrow 10.83$} & $82.10$ \scriptsize{$\downarrow 11.81$}\\
    PLAT * \cite{wang2020enresnet} & $94.73$ & $88.28$ \scriptsize{$\downarrow 6.45$} & $88.56$\scriptsize{$\downarrow 6.17$} \\
    \midrule
    \multicolumn{4}{c}{\textbf{Noise Injection}} \\
    RSE *\cite{liu2018towards} & $95.59$ & $77.86$ \scriptsize{$\downarrow 17.73$} & N/A \\
    ENResNet *\cite{wang2020enresnet} & $83.33$ & $74.34$ \scriptsize{$\downarrow 8.99$} & N/A \\
    \midrule
    \multicolumn{4}{c}{\textbf{Data Augment}} \\
    AugMix § \cite{hendrycks2019augmix} & $95.83$ & $89.09$ \scriptsize{$\downarrow 6.74$} & $88.71$ \scriptsize{$\downarrow 7.12$} \\
    AutoAug *\cite{cubuk2019autoaugment} & $95.61$ & $85.37$ \scriptsize{$\downarrow 10.24$} & $76.47$\scriptsize{$\downarrow 19.14$}  \\
    \midrule
    \multicolumn{4}{c}{\textbf{Generative Methods}} \\
    SBGC  \cite{zimmermann2021score} & $95.04$ & $76.24$ \scriptsize{$\downarrow 18.80$} & $75.38$\scriptsize{$\downarrow 19.66$}  \\
    PDE+  \cite{yuan2024pde} & $95.59$ & $89.11$ \scriptsize{$\downarrow 6.48$} & $85.59$\scriptsize{$\downarrow 10.00$} \\
    \hdashline
    \multicolumn{4}{c}{\textbf{Our methods}} \\
    PRG-GVP-S  & $97.35$ & $91.21$ \scriptsize{$\downarrow 6.14$} & $92.17$ \scriptsize{$\downarrow 5.18$} \\
    PRG-ICFM-S  & $97.59$ & $92.13$ \scriptsize{$\downarrow 5.46$} & $93.07$ \scriptsize{$\downarrow 4.52$} \\
    PRG-OTCFM-S  & $97.65$ & $92.26$ \scriptsize{$\downarrow 5.39$} & $93.10$ \scriptsize{$\downarrow 4.55$} \\
    \bottomrule
  \end{tabular}
 \caption{\textbf{(OOD: Image Corruptions)} Comparison on CIFAR-10-C, including Noise, Blur, Weather, and Digital corruptions. Results are referenced from original papers or \cite{yuan2024pde, zimmermann2021score}. * denotes ResNet-18 as the base model, while § indicates ResNeXt-29.}
  \label{tab:oodcifar}
\end{table}

\begin{table}[h]
  \centering
  \small
  \begin{tabular}{@{}lccc@{}}
    \toprule
    \textbf{Method} & \textbf{Param. (M)} & \textbf{Cifar Acc.} & \textbf{Tiny Acc.} \\
    \midrule
    \multicolumn{4}{c}{\textbf{Classic self-supervised learning methods}} \\
    MAE (ViT-B/16) \cite{liu2023good} & $86$ & $96.5$ & $76.5$ \\
    SimCLR Res-50-4x \cite{chen2020simple} & $375$ & $98.6$ & N/A \\
    \midrule
    \multicolumn{4}{c}{\textbf{Generative methods}} \\
    DDAE-DiT-XL2 \cite{xiang2023denoising} & $338$ & $98.4$ & $77.8$ \\
    iGPT-L \cite{chen2020generative} & $1362$ & $99.0$ & N/A \\
    \hdashline
    \multicolumn{4}{c}{\textbf{Our methods}} \\
    PRG-SiT-XL2 & $338$ & $98.72$ & $78.33$ \\
    \bottomrule
  \end{tabular}
  \caption{Comparison of models pretrained on ImageNet-1k and transferred (fine-tuning) to CIFAR-10 and Tiny-ImageNet.}
  \label{tab:performance_comparison}
\end{table}

\subsubsection{Transferring Features}
\label{sec:trans}
A main goal of the pre-training/fine-tuning paradigm is to learn transferrable features.
We hypothesize that advancements like powerful pretrained models in the generative model community can contribute to our method.
To validate this, we conducted experiments by transferring the pretrained SiT-XL~\cite{ma2024sit} model to two other datasets.
Since SiT provides only class-conditional checkpoints, 
we adopted an unconditional approach by setting the label parameter to null. 
Specifically, we fine-tuned the model for 28 epochs on the CIFAR-10 dataset and for 45 epochs on the Tiny-ImageNet dataset.
As demonstrated in \cref{tab:performance_comparison}, our method shows superior transfer learning performance.
The results suggest that our algorithm benefits from larger datasets and the latent space of generative models.

\subsection{Ablation Studies}
In this section, we conduct three ablation studies to further evaluate the applicability of our method. Additional experiments, including analyses of loss type, ODE solver choice, dual-task supervision, and $t_{\text{span}}$, are provided in \cref{sec:Ablation Studies appendix}.

\subsubsection{Generative Model Type}
We explored various generative model types, particularly focusing on path selection.
To assess the impact of these paths on our method, we conducted experiments with three model types on the CIFAR-10 and Tiny-ImageNet datasets.
Following \cite{liu2022flow}, we quantified the straightness of various continuously differentiable trajectories.
% demonstrates how straightness varies across different generation approaches during fine-tuning. 
Interestingly, while all three model variants achieved comparable performance (\cref{tab:generative_model_comparison}),
models with higher straightness required fewer pre-training steps to attain equivalent performance levels and demonstrated potential for superior outcomes.
\begin{table}[h]
  \centering
  \small
  \begin{tabular}{@{}lccc@{}}
    \toprule
    \textbf{Generative Model Type} & \textbf{GVP} & \textbf{ICFM} & \textbf{OTCFM} \\
    \midrule
    CIFAR-10 (Accuracy) & $97.35$ & $97.59$ & $97.65$ \\
    First to $97$\% (Epoch) & $162$ & $135$ & $128$ \\
    Straightness & $7.36$ & $0.34$ & $0.11$ \\
    \midrule
    Tiny-ImageNet (Accuracy) & $70.98$ & $71.12$ & $71.33$ \\
    First to $70$\% (Epoch) & $157$ & $125$ & $106$ \\
    Straightness & $6.25$ & $0.54$ & $0.15$ \\
    \bottomrule
  \end{tabular}
  \caption{\textbf{(Generative Model Type)} Performance of different generative model types on the CIFAR-10 and Tiny-ImageNet datasets.}
  \label{tab:generative_model_comparison}

\end{table}

\subsubsection{Effect of $\beta$ on Accuracy}
During pre-training, we introduced $\beta$ in \cref{eq:cross_entropy_loss} to enhance the mutual information lower bound while balancing the downstream task objective. \Cref{tab:balance} shows PRG scores across datasets for different $\beta$ values.
\begin{table}[bth]
  \centering
  \small
  % \fbox{\rule{0pt}{0.5in} \rule{0.9\linewidth}{0pt}}
  \setlength{\abovecaptionskip}{2pt}
  \begin{tabular}{@{}lcccc@{}}
    \toprule
    \textbf{$\beta$} & \textbf{1} & \textbf{10} & \textbf{100} \\
    \midrule
    Acc. (CIFAR/Tiny) & $0.973$/$0.712$ & $0.974$/$0.711$ & $0.965$/$0.673$ \\
    \bottomrule
  \end{tabular}
\caption{\textbf{($\beta$ values)} Accuracy scores of PRG with varying $\beta$ values on the CIFAR-10 and Tiny-ImageNet datasets.}
  \label{tab:balance}

\end{table}

\subsubsection{Scaling Up Network Parameters}
We investigated the impact of scaling up U-Net models of varying network parameter sizes~\cite{nichol2021improved}.
\Cref{tab:model_comparison} provides a comparative analysis following fine-tuning on the CIFAR-10 and Tiny-ImageNet datasets.
The findings suggest that increasing the model size results in a modest enhancement of final classification accuracy.

\begin{table}[h]
  \centering
  \small
  \begin{tabular}{@{}lcccc@{}}
    \toprule
    \textbf{Model} & \textbf{Param (M)} & \textbf{CIFAR Acc.} & \textbf{Tiny Acc.} \\
    \midrule
    PRG-ICFM-B & $32$ & $97.35$ & $70.94$ \\
    PRG-ICFM-S & $42$ & $97.59$ & $71.12$ \\
    PRG-ICFM-L & $66$ & $97.75$ & $71.30$ \\
    PRG-ICFM-XL & $122$ & $97.88$ & $71.80$ \\
    \bottomrule
  \end{tabular}
  \caption{\textbf{(Scaling Capability)} Performance gains of increasing U-Net model sizes on the CIFAR-10 and Tiny-ImageNet datasets.}
  \label{tab:model_comparison}
\end{table}

%% file: sec/6_conclusion.tex
\section{Conclusion}
This paper proposes to turn a pretrained continuous-time flow/diffusion generator upside-down: running the model backward produces multi-level features that, after light fine-tuning, serve as an unsupervised representation extractor.
We systematically investigate the necessary designs of the two-stage pre-training and fine-tuning paradigm. 
Theoretical analysis supports the efficacy of the first-stage generative pre-training, while empirical verifications provide insights for designing the fine-tuning pipeline. 
Leveraging these simple yet effect techniques and findings, our method achieves state-of-the-art performance in image classification tasks using generative models. Additional experiments, such as those on OOD problems and transfer learning scenarios, further validate the robustness and other advantages of our approach. 
However, there remains room for improvement, including the integration of large foundation pretrained generative models from the open-source community and optimizing the training epoch/speed for faster adaptation to downstream tasks.

%% file: sec/X_suppl.tex
\clearpage
\maketitlesupplementary
\appendix

\setcounter{table}{0} 
\setcounter{equation}{0}
\setcounter{figure}{0}

\section{Why PRG is effective?}
\label{sec:myadvantage}
What a model can generate, it can understand: pixel-perfect reconstruction means the visual breadcrumbs are all there.
It just needs a bit of fine-tuning to tie the generative pathway to the labels.
Raw-image baselines often fall prey to "shortcut learning".
Because neighboring feature pixels have heavily overlapping receptive fields, the network may rely on trivial cues and end up with weak representations.
In contrast, PRG reuses the pretrained reverse generative pathway, recycling each ODE-solver output as the input for the next iteration.
This loop drives feature extraction at a semantic level without losing low-level detail and the injected noise perturbations make the representations significantly more robust. \cref{sec:my432} show that every point along the trajectory possesses strong representational capacity, and the OOD experiments also provide compelling evidence for this claim.

In the first pre-training stage, since no downstream task information is available, it remains unclear which features are most relevant. Consequently, no compression is performed during pre-training. 
Instead, the model aims to approximate the optimal representation as closely as possible in the absence of downstream information by leveraging unconditional flow matching to learn representations, effectively maximizing the lower bound of mutual information.

According to the manifold assumption \cite[Chapter 20]{murphy2023probabilistic}, data lies on a low-dimensional manifold $\mathcal{M}$ of intrinsic dimension $d^*$, which is significantly smaller than the ambient dimension $D$. Although, in theory, intermediate latents encode the same information as the original data, using these latents as inputs for downstream tasks is more meaningful. A well-trained flow model effectively extracts data from the low-dimensional manifold $\mathcal{M}$ and lifts it to the ambient space $\mathbb{R}^D$, ensuring that every sample in $\mathbb{R}^D$ remains semantically meaningful. In contrast, directly sampling from the original data space does not necessarily preserve semantic coherence. For instance, generating a $64 \times 64$ image by sampling from a $64 \times 64$ multivariate Gaussian distribution would typically result in a meaningless noise-like image resembling a snowflake pattern.

Moreover, since the flow trajectories of an ordinary differential equation (ODE) do not intersect~\cite{Chen2018}, points within a given region remain confined, thereby preserving topological relationships throughout the transformation process.

Subsequent fine-tuning for downstream tasks is equally crucial. By adopting techniques such as optimal transport matching and gradient guidance, the model undergoes efficient adaptation, selectively discarding redundant information while selecting and enhancing task-critical features. As shown in Figure~\ref{fig:logp-acc}, mutual information decreases during fine-tuning, whereas task accuracy improves.

\subsection{What’s the difference between fine-tuning and training a classifier based on the ODE architecture? }
\label{sec:difference}
The effectiveness of fine-tuning depends significantly on whether the model undergoes pre-training beforehand. If a classifier is trained directly using the ODE architecture without pre-training, the model must learn meaningful intermediate latents solely through a supervised loss. This approach often fails, as the model struggles to discover good latent representations in the absence of prior knowledge. Our experiments~(\cref{sec:without pre-training}) also confirm that this method performs poorly.

In contrast, pre-training allows the model to extract a rich set of useful features as intermediate latents. During fine-tuning, the model can then selectively retain and enhance features relevant to downstream tasks while compressing redundant information in the latent space. This process leads to a more effective and structured representation, ultimately improving downstream task performance.

\subsubsection{Comparison with Generation without pre-training}
\label{sec:without pre-training}
We compare PRG with classifiers trained without generative pre-training. To ensure fairness, we train the latter model for 600 epochs until its performance no longer improves. As shown in \cref{tab:classification_performance}, PRG without pre-training, relying solely on a single supervision signal, may discard useful information, leading to suboptimal performance. This highlights that the intermediate latent variables of a pretrained flow model provide a strong initialization, requiring only slight fine-tuning for optimal feature extraction.
\begin{table}[bth]
  \centering
  \small
  \begin{tabular}{@{}lccc@{}}
    \toprule
    \textbf{} & \textbf{CIFAR-10} & \textbf{Tiny-ImageNet} & \textbf{ImageNet} \\
    \midrule
    PRG w/o pre-training & $0.70$ & $0.35$ & $0.42$ \\
    PRG & $0.97$ & $0.71$ & $0.78$ \\
    \bottomrule
  \end{tabular}
\caption{\textbf{Effect of pre-training:} Performance comparison of PRG with and without pre-training on the CIFAR-10 and Tiny-ImageNet.}
  \label{tab:classification_performance}
\end{table}

\subsection{Model-agnosticity}
Our method follows the model-agnostic principle as defined in Appendix D.5 of \cite{song2020score}, where different architectures achieve comparable encoding quality through flow matching. This indicates that architectural constraints stem from implementation choices rather than the framework itself.

Notably, while early diffusion models primarily relied on U-Nets for practical stability, recent work \cite{bao2023all} has demonstrated that flow matching can be successfully achieved with minimal architectural requirements, such as shallow skip connections. In this work, we achieve state-of-the-art performance across both U-Net~(\cref{sec:unet}) and Transformer~(\cref{sec:trans}) architectures, further validating the generality of our approach.

\subsection{Infinite-Layer Expressiveness}
The infinite-layer extractor boosts accuracy by extending training rather than inference steps (Fig.6: CIFAR 0.46 vs. 0.97, Tiny 0.19 vs. 0.71). However, gains plateau after a certain point, with more complex datasets requiring longer training for optimal performance. Meanwhile, the optimal flow-matching setup achieves full experimental coverage in just $5$ inference steps. \textbf{Full evaluation takes 3 seconds for CIFAR and 9 s for the Tiny-ImageNet testing set.}

% \section{Model-agnosticity}
% Our method adheres to model-agnosticism as defined in (\textit{\textcolor[HTML]{228B22}{score-based generative modeling through stochastic differential equations  Appendix D.5}}), where different architectures achieve comparable encoding quality through flow matching. 
% This shows that architectural constraints come from implementation choices, not the framework itself. While early diffusion models used U-Nets for practical stability, 
% recent work (\textit{\textcolor[HTML]{228B22}{U-ViT}}) demonstrates successful flow matching with minimal architectural requirements (shallow skip connections). 

\section{Additional Implementation Details}
\label{sec:Additional Implementation Details}
\subsection{Training Process Details}
\label{sec:Training Process Details}
\begin{table}[h]
\centering
\small
\begin{tabular}{@{}lccc@{}}
\toprule
\textbf{Dataset} & \textbf{CIFAR-10} & \textbf{Tiny-ImageNet} & \textbf{ImageNet} \\ 
\midrule
\textbf{GPU} & 8$\times$A100 & 8$\times$A100 & 8$\times$A100 \\ 
\textbf{Optimizer} & Adam & Adam & Adam \\ 
\textbf{LR base} & 1e-4 & 1e-4 & 1e-4 \\ 
\textbf{Epochs} & 1000 & 1000 & 2000 \\ 
\textbf{Batch Size} & 256 & 128 & 128 \\ 
\bottomrule
\end{tabular}
\caption{Experimental settings across datasets for \textbf{pre-training}.}
\label{tab:experimental_settings_pretrain}
\end{table}

\begin{table}[h]
\centering
\small
\begin{tabular}{@{}lccc@{}}
\toprule
\textbf{Dataset} & \textbf{CIFAR-10} & \textbf{Tiny-ImageNet} & \textbf{ImageNet} \\ 
\midrule
\textbf{GPU} & 4$\times$A100 & 8$\times$A100 & 32$\times$A100 \\ 
\midrule
\textbf{Optimizer} & AdamW & AdamW & AdamW \\ 
\textbf{Eps} & 1e-8 & 1e-8 & 1e-8 \\ 
\textbf{Betas} & (0.9, 0.999) & (0.9, 0.999) & (0.9, 0.999) \\ 
\textbf{LR base} & 1.25e-4 & 1.25e-4 & 1.25e-4 \\ 
\textbf{Weight Decay} & 0.05 & 0.05 & 0.05 \\ 
\midrule
\textbf{Scheduler} & CosineLR & CosineLR & CosineLR \\ 
\textbf{Warmup LR base} & 1.25e-7 & 1.25e-7 & 1.25e-7 \\ 
\textbf{Min LR base} & 1.25e-6 & 1.25e-6 & 1.25e-6 \\ 
\textbf{Epochs} & 200 & 200 & 200 \\ 
\textbf{Warmup Epochs} & 5 & 10 & 10 \\ 
\midrule
\textbf{Image Size} & 32 & 64 & 64 \\ 
\textbf{Batch Size} & 256 & 128 & 128 \\ 
\textbf{T Span} & 20 & 32 & 64 \\ 
\textbf{Solver} & Euler& Euler& Euler \\ 
\bottomrule
\end{tabular}
\caption{Experimental settings across datasets for \textbf{fine-tuning}.}
\label{tab:experimental_settings_finetune}
\end{table}

To enhance the reproducibility of results across various multi-stage and multi-GPU experiments, we calculate the learning rate using \cref{eq:learning_rates}.
\begin{equation}
\small
\label{eq:learning_rates}
\begin{aligned}
\text{LR} &= \text{LR}_{\text{base}} \times \frac{\text{num processes} \times \text{Batch Size}}{512} \\
\text{Warmup LR} &= \text{Warmup LR}_{\text{base}} \times \frac{\text{num processes} \times \text{Batch Size}}{512} \\
\text{Min LR} &= \text{Min\_LR}_{\text{base}} \times \frac{\text{num processes} \times \text{Batch Size}}{512}
\end{aligned}
\end{equation}

\subsection{Evaluation of Training Efficiency}
\label{sec:Evaluation of Training Efficiency}
Most research experiments, including the main experiments and ablation studies, are completed within several hours.
To demonstrate training efficiency concretely, \cref{tab:training_times} reports the run-time per epoch for each data set used in our experiments.
\begin{table}[h]
  \centering
  \small
  \begin{tabular}{@{}lccc@{}}
    \toprule
    \textbf{$t_\text{span}/t_\text{cutoff}$} & \textbf{CIFAR-10} & \textbf{TinyImageNet} & \textbf{ImageNet*} \\
    \midrule
    $20/5$  & $49\text{s}$   & $307\text{s}$   & N/A \\
    $20/10$ & $103\text{s}$ & $681\text{s}$ & $2220\text{s}$ \\
    $20$    & $187\text{s}$ & $1437\text{s}$ & $3480\text{s}$ \\
    $32/16$ & $170\text{s}$ & $1137\text{s}$ & N/A \\
    \bottomrule
  \end{tabular}
  \caption{Mean time cost per epoch during the training process. * means that the model is trained on $4*8$ A100 GPUs. $t_\text{span}$ represents the total sampling length, while $t_\text{cutoff}$ indicates the point along the trajectory where fine-tuning begins. For example, $20/10$ means a trajectory spanning $20$ steps from $x_0$ to $x_1$, with fine-tuning starting from the midpoint of the trajectory.}
  \label{tab:training_times}
\end{table}

\subsection{Evaluation of Inference Efficiency}
\label{sec:evaluation_inference_efficiency}
\Cref{tab:inference_times} presents the inference efficiency of our method on the CIFAR-10 and Tiny-ImageNet test sets.

\begin{table}[h]
    \centering
    \small
    \begin{tabular}{@{}lcc@{}}
        \toprule
        \textbf{Model} & \textbf{CIFAR-10 (s)} & \textbf{Tiny-ImageNet (s)} \\
        \midrule
        PRG-GVP-S   & 4  & 10  \\
        PRG-ICFM-S  & 3  & 9   \\
        PRG-OTCFM-S & 3  & 8   \\
        \bottomrule
    \end{tabular}
    \caption{Mean inference time per epoch on the CIFAR-10 and Tiny-ImageNet test datasets.}
    \label{tab:inference_times}
\end{table}

\section{Ablation Studies}
\label{sec:Ablation Studies appendix}

\subsection{Loss Type}
\cref{tab:loss_comparison} shows the results of different loss types. Compared to the standard cross entropy loss, label smoothing reduces overconfident predictions, improves model calibration, and improves robustness.
\begin{table}[bht]
  \centering
  \small
  \begin{tabular}{@{}lcc@{}}
    \toprule
    \textbf{Dataset} & \textbf{LabelSmooth Loss} & \textbf{Cross-Entropy Loss} \\
    \midrule
    CIFAR-$10$ & $97.59$ & $96.18$ \\
    TinyImageNet & $71.12$ & $70.15$ \\
    \bottomrule
  \end{tabular}
  \caption{\textbf{(Loss Type)} Comparison of LabelSmooth Loss and Cross-Entropy Loss on different datasets.}
  \label{tab:loss_comparison}
\end{table}

\subsection{ODE Solver Type}
During fine-tuning, we evaluated different ODE solvers for the reverse process: Euler (first-order), Midpoint (second-order via midpoint evaluations), RK4 (fourth-order Runge-Kutta), and Dopri5 (adaptive step sizes with a fifth-order method). \cref{tab:solver_performance on cifar-10,tab:solver_performance on Tiny-ImageNet} compares their performance on the Cifar-10, Tiny-ImageNet dataset. The results show no significant performance differences, underscoring the method’s consistent effectiveness across various solvers.
\begin{table}[bht]
  \centering
  \small
  \begin{tabular}{@{}lcccc@{}}
    \toprule
    \textbf{solver\_type} & \textbf{Euler} & \textbf{Midpoint} & \textbf{RK4} & \textbf{Dopri5} \\
    \midrule
    OTCFM & $97.65$ & $97.63$ & $97.66$ & $97.72$ \\
    ICFM & $97.59$ & $97.55$ & $97.56$ & $97.61$ \\
    GVP & $97.35$ & $97.30$ & $97.32$ & $97.41$ \\
    \bottomrule
  \end{tabular}
  \caption{\textbf{(ODE Solver)} Performance of various solvers on Cifar-10. Different solvers don't yield obvious differences.}
  \label{tab:solver_performance on cifar-10}
\end{table}

\begin{table}[bht]
  \centering
  \small
  \begin{tabular}{@{}lcccc@{}}
    \toprule
    \textbf{Solver Type} & \textbf{Euler} & \textbf{Midpoint} & \textbf{RK4} & \textbf{Dopri5} \\
    \midrule
    PRG-OTCFM & $71.33$ & $71.29$ & $71.30$ & $71.36$ \\
    PRG-ICFM & $71.12$ & $71.11$ & $71.13$ & $71.23$ \\
    PRG-GVP & $70.89$ & $70.99$ & $70.84$ & $70.85$ \\
    \bottomrule
  \end{tabular}
  \caption{\textbf{(ODE Solver)} Performance of various solvers on Tiny-ImageNet. Different solvers don't yield obvious differences.}
  \label{tab:solver_performance on Tiny-ImageNet}
\end{table}

\subsection{Details of Out-of-Distribution Experiments}
\label{sec:Details of Out-of-Distribution Experiments}
\begin{table}[bth]
  \centering
  \small
  \begin{tabular}{@{}lccc@{}}
  \toprule
  \multicolumn{4}{c}{\textbf{Tiny-ImageNet-C}} \\
  \midrule
  \textbf{Method} & \textbf{Clean} & \textbf{Average} & \textbf{Corruption-$5$} \\
  \midrule
  \textbf{Adversarial Training} & & & \\
  PGD \cite{madry2017towards} & $51.08$ & $33.46$ \scriptsize{$\downarrow 17.62$} & $24.00$ \scriptsize{$\downarrow 27.08$} \\
  PLAT \cite{kireev2022effectiveness} & $51.29$ & $37.92$ \scriptsize{$\downarrow 13.37$} & $29.05$ \scriptsize{$\downarrow 22.24$} \\
  \midrule
  \textbf{Noise Injection} & & & \\
  RSE \cite{liu2018towards} & $53.74$ & $27.99$ \scriptsize{$\downarrow 25.75$} & $18.92$ \scriptsize{$\downarrow 34.82$} \\
  ENResNet \cite{wang2020enresnet} & $49.26$ & $25.83$ \scriptsize{$\downarrow 23.43$} & $19.01$ \scriptsize{$\downarrow 30.25$} \\
  \midrule
  \textbf{Data Augment} & & & \\
  AugMix \cite{hendrycks2019augmix} & $52.82$ & $37.74$ \scriptsize{$\downarrow 15.08$} & $28.66$ \scriptsize{$\downarrow 24.16$} \\
  AutoAug \cite{cubuk2019autoaugment} & $52.63$ & $35.14$ \scriptsize{$\downarrow 17.49$} & $25.36$ \scriptsize{$\downarrow 27.27$} \\
  \midrule
  \textbf{Generative Methods} & & & \\
  PDE+ \cite{yuan2024pde} & $53.72$ & $39.41$ \scriptsize{$\downarrow 14.31$} & $30.32$ \scriptsize{$\downarrow 23.40$} \\
  \hdashline
  PRG-ICFM-S (ours) & $56.85$ & $46.93$ \scriptsize{$\downarrow 9.92$} & $33.32$ \scriptsize{$\downarrow 23.53$} \\
  \bottomrule
  \end{tabular}
  \caption{\textbf{(OOD: extrapolated datasets)} Performance on Tiny-ImageNet-C.Averge represents the accuracy across all corruption levels, with corruption severity ranging from $1$ to $5$.}
  \label{tab:oodTinyImageNet}
\end{table}
There is no direct correspondence between the test images of Tiny ImageNet and Tiny ImageNet-C, and the images in Tiny ImageNet-C do not overlap with the training images of Tiny ImageNet. We report the comparative results on Tiny ImageNet-C in \cref{tab:oodTinyImageNet}.

\subsection{The Number of Timesteps}
Our findings in \cref{tab:t_span_analyse} show that longer time spans generally lead to better accuracy. 
On CIFAR-10 with the ICFM flow model, a $t$-span of 10 achieves accuracy comparable to the best result at $t=100$. 
In contrast, TinyImageNet requires a $t$-span of 15 to achieve similar performance.

\begin{table}[h]
  \centering
  \small
  \begin{tabular}{@{}lccccccccc@{}}
    \toprule
    \textbf{T-span} & $\mathbf{2}$  & $\mathbf{5}$ & $\mathbf{10}$  & $\mathbf{50}$ & $\mathbf{100}$ \\
    \midrule
    GVP CIFAR-10 & $30.54$   & $90.23$ & $93.26$ & $97.50$ & $97.55$ \\
    GVP Tiny ImageNet & $5.06$  & $48.95$ & $53.24$  & $71.05$ & $71.18$ \\
    ICFM CIFAR-10 & $31.18$  & $92.35$ & $97.02$  & $97.60$ & $97.61$ \\
    ICFM Tiny ImageNet & $6.01$ & $60.06$ & $65.16$  & $71.20$ & $71.58$ \\
    \bottomrule
  \end{tabular}
  \caption{Comparison of Performance Over Varying Time Spans.}
  \label{tab:t_span_analyse}
\end{table}

\begin{figure*} [bth]
  \centering
  \small
  \includegraphics[width=\linewidth]{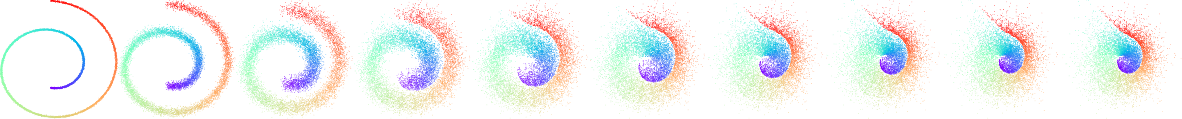}
  \caption{Reverse Generation Process on the Swiss Roll Dataset. Each color represents a different class. After diffusion, samples from the same class become more clustered, while the previously unoccupied white space, corresponding to out-of-class regions, is pushed outward.}
  \label{fig:appendix_big_picture}
\end{figure*}

\begin{figure*} [bth]
  \centering
  \small
  \includegraphics[width=\linewidth]{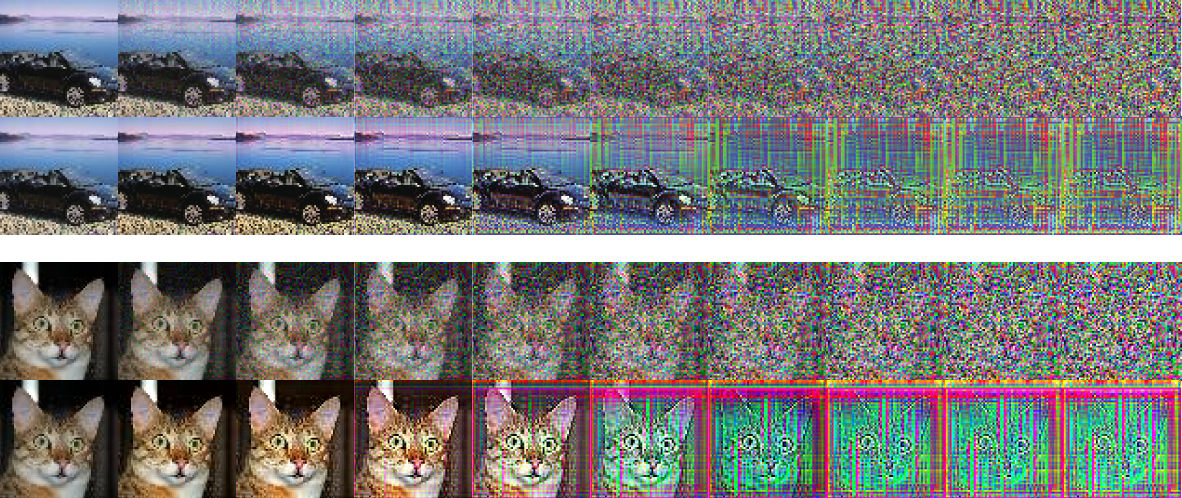}
  \caption{Reverse Generation Process on the TinyImageNet val set. The first row represents the fully pretrained reverse generative process,
the second row shows the reverse generative process after extensive fine-tuning.}
  \label{fig:val-set}
\end{figure*}

\section{Reverse Generation Process}
\label{sec:Reverse Generation Process}
\begin{figure*} [bth]% 'b' option places the figure at the bottom of the page
  \centering
  \small
  \includegraphics[width=\linewidth]{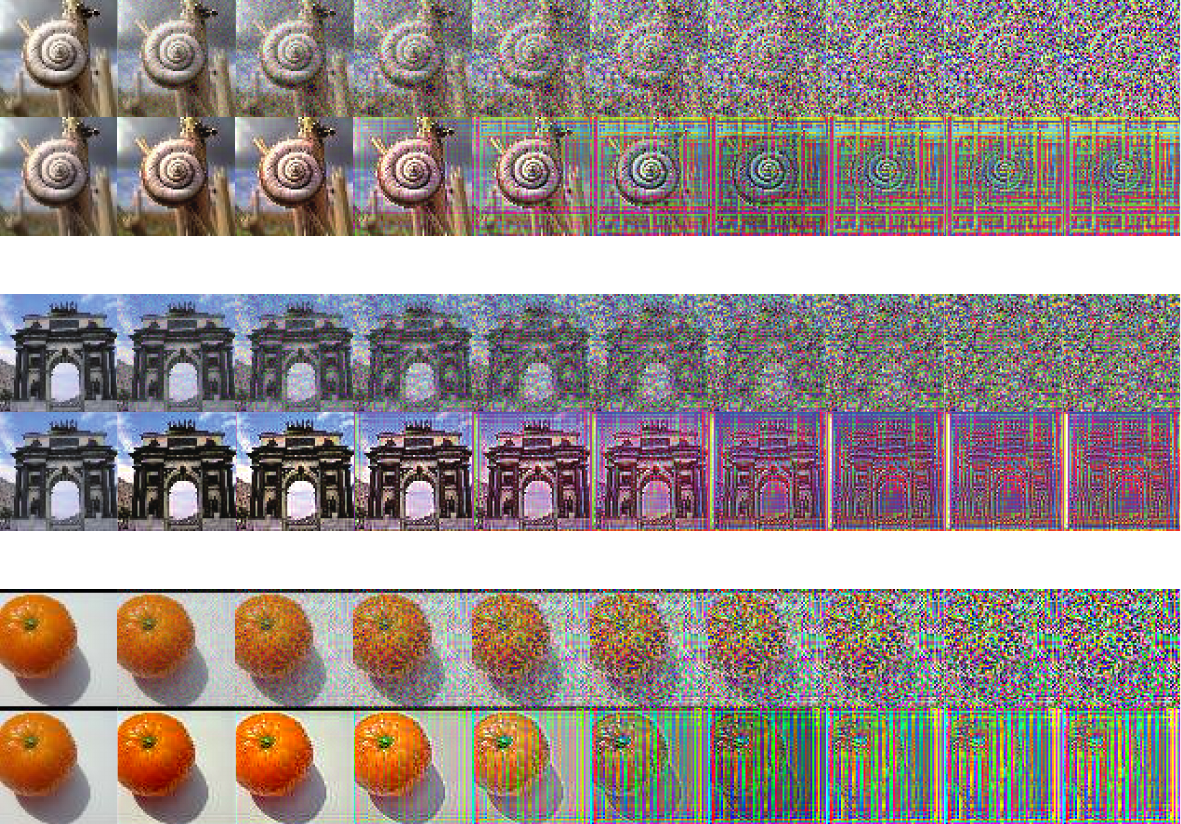}
  \caption{Reverse Generation Process on the TinyImageNet train set. The first row represents the fully pretrained reverse generative process, the second row shows the reverse generative process after extensive fine-tuning.}
  \label{fig:train_set}
\end{figure*}

\begin{figure*}[bth] % 'b' option places the figure at the bottom of the page
  \centering
  \small
  \includegraphics[width=\linewidth]{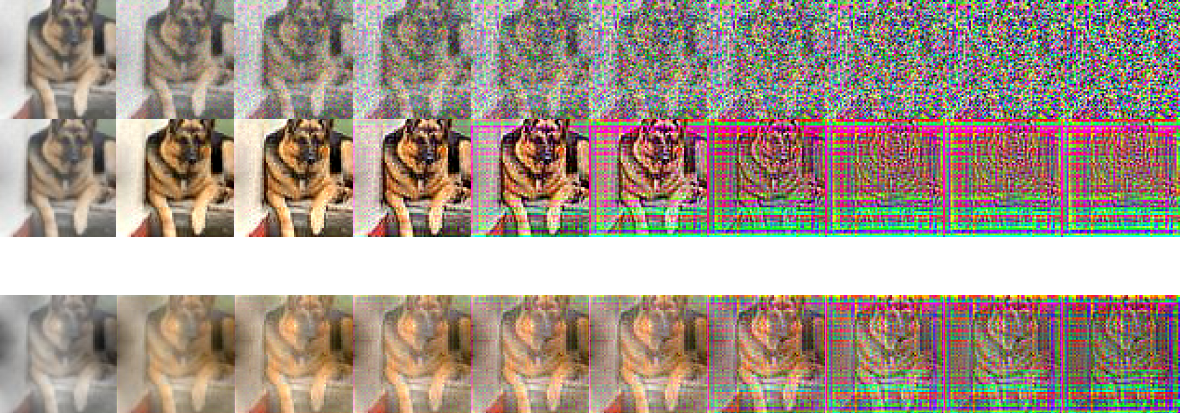}
\caption{Reverse Generation Process on the TinyImageNet-C Dataset. The first row represents the fully pretrained reverse generative process, the second row shows the reverse process after extensive fine-tuning, and the third row illustrates the reverse generative process under fog corruption.}
  \label{fig:c-set}
\end{figure*}

Fig. \ref{fig:appendix_big_picture} illustrates the reverse generation process from \(x_0\) to \(x_1\) after fine-tuning. Furthermore, Figs. \ref{fig:val-set} and \ref{fig:train_set} present the reverse generation results on the TinyImageNet dataset after pre-training and fine-tuning, respectively. Finally, Fig. \ref{fig:c-set} demonstrates the reverse process before and after applying fog corruption to the images.

% \begin{figure}
%   \centering
%   \small
%   \includegraphics[width=\linewidth]{img/cifar_straightness.pdf}
%   \caption{Strightness of the Pretrain on CIFAR-10.}
%   \label{fig:strightness-cifar}
% \end{figure}

% \begin{table*}
%   \centering
%   \small
%   \begin{tabular}{lcccccccccc}
%     \toprule
%     & \multicolumn{1}{c}{\textbf{Tiny-ImageNet}} & \multicolumn{8}{c}{\textbf{Tiny-ImageNet-C}} \\
%     \cmidrule(lr){2-2} \cmidrule(lr){3-10}
%     \textbf{Model} & \textbf{Acc.} & \textbf{Brightness} & \textbf{Contrast} & \textbf{Defocus\_blur} & \textbf{Fog} & \textbf{Gaussian\_blur} & \textbf{Pixelate} & \textbf{Snow} & \textbf{Speckle\_noise} \\
%     \midrule
%     gvp-model & $70.98$ & $56.18$ & $53.14$ & $53.48$ & $53.69$ & $53.17$ & $48.99$ & $52.38$ & $56.78$ \\
%     ot-model & $71.33$ & $56.54$ & $53.33$ & $53.68$ & $53.59$ & $54.27$ & $49.66$ & $52.26$ & $56.88$ \\
%     ICFM-model & $71.12$ & $56.30$ & $52.54$ & $53.33$ & $53.80$ & $54.17$ & $49.54$ & $52.16$ & $56.25$ \\
%     \bottomrule
%   \end{tabular}
%   \caption{Performance on Tiny ImageNet-C.}
%   \label{tab:oodTiny-ImageNet}
% \end{table*} 
% \section{Detailed Information of the Pretrained Reversible Generation}
% \section{Derivations of Formulas in the Paper}

\vfill  % This command pushes the content to the bottom of the page

